\documentclass{article}

\usepackage{subcaption}
\usepackage{diagbox}
\usepackage{multirow}
\usepackage{multicol}
\usepackage{algorithm}
\usepackage{algorithmic}
\usepackage{amsmath}

\DeclareMathOperator*{\argmin}{arg\,min}
\newtheorem{theorem}{Observation}
\usepackage[firstpage]{draftwatermark}
\SetWatermarkText{
 \hspace*{3.5in}
 \raisebox{10.0in}{
  \includegraphics[height=0.9in]{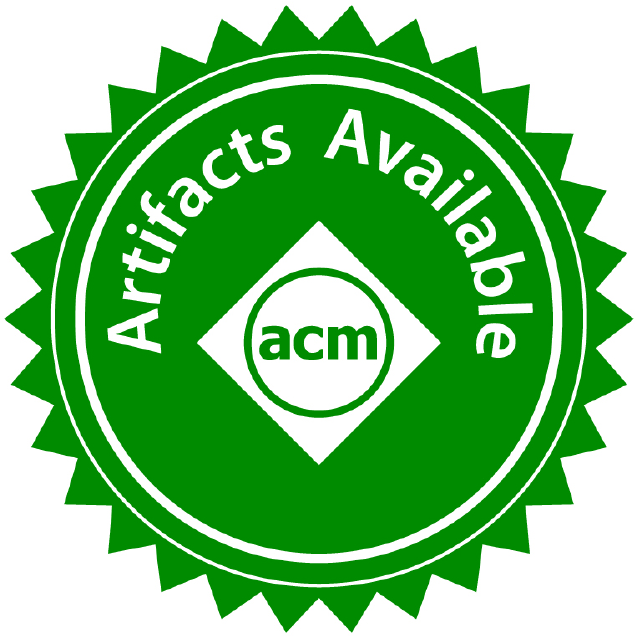}
  \includegraphics[height=0.9in]{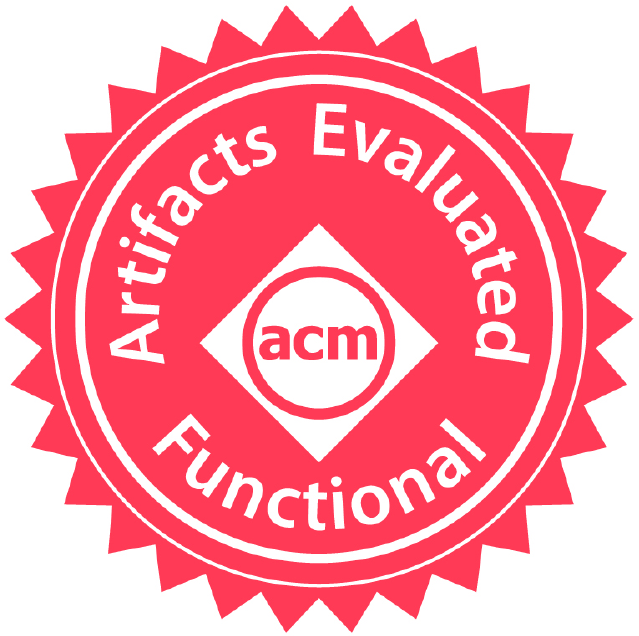}
 }
}
\SetWatermarkAngle{0}

\usepackage{microtype}
\usepackage{graphicx}
\usepackage{booktabs} 

\usepackage{hyperref}

\usepackage[accepted]{mlsys2023}

\mlsystitlerunning{Pre-train and Search: Efficient Embedding Table Sharding with Pre-trained Neural Cost Models}

\begin{document}

\twocolumn[
\mlsystitle{Pre-train and Search: Efficient Embedding Table Sharding with Pre-trained Neural Cost Models}




\begin{mlsysauthorlist}
\mlsysauthor{Daochen Zha}{rice}
\mlsysauthor{Louis Feng}{meta}
\mlsysauthor{Liang Luo}{meta}
\mlsysauthor{Bhargav Bhushanam}{meta}
\mlsysauthor{Zirui Liu}{rice}
\mlsysauthor{Yusuo Hu}{meta}
\mlsysauthor{Jade Nie}{meta}
\mlsysauthor{Yuzhen Huang}{meta}
\mlsysauthor{Yuandong Tian}{meta}
\mlsysauthor{Arun Kejariwal}{meta}
\mlsysauthor{Xia Hu}{rice}
\end{mlsysauthorlist}

\mlsysaffiliation{rice}{Department of Computer Science, Rice University, USA}
\mlsysaffiliation{meta}{Meta Platforms, Inc., USA}

\mlsyscorrespondingauthor{Daochen Zha}{daochen.zha@rice.edu}

\mlsyskeywords{Embedding Table, Recommendation Models, Pre-training, Distributed System, ML for Systems}

\vskip 0.3in

\begin{abstract}
Sharding a large machine learning model across multiple devices to balance the costs is important in distributed training. This is challenging because partitioning is NP-hard, and estimating the costs accurately and efficiently is difficult. In this work, we explore a \emph{``pre-train, and search''} paradigm for efficient sharding. The idea is to pre-train a universal and once-for-all neural network to predict the costs of all the possible shards, which serves as an efficient sharding simulator. Built upon this pre-trained cost model, we then perform an online search to identify the best sharding plans given any specific sharding task. We instantiate this idea in deep learning recommendation models (DLRMs) and propose NeuroShard for embedding table sharding. NeuroShard pre-trains neural cost models on augmented tables to cover various sharding scenarios. Then it identifies the best column-wise and table-wise sharding plans with beam search and greedy grid search, respectively. Experiments show that NeuroShard significantly and consistently outperforms the state-of-the-art on the benchmark sharding dataset, achieving up to 23.8\% improvement. When deployed in an ultra-large production DLRM with multi-terabyte embedding tables, NeuroShard achieves 11.6\% improvement in embedding costs over the state-of-the-art, which translates to 6.6\% end-to-end training throughput improvement. To facilitate future research of the \emph{``pre-train, and search''} paradigm in ML for Systems, we open-source our code at \url{https://github.com/daochenzha/neuroshard}

\end{abstract}
]



\printAffiliationsAndNotice{} 

\section{Introduction}
Deep learning recommendation models (DLRMs) are one of the most important machine learning applications~\cite{zhang2019deep,cheng2016wide,naumov2019deep,tan2023bring,zhouinterest}. For example, DLRMs account for more than 50\% of training and 80\% inference demands in Meta~\cite{naumov2020deep,gupta2020architectural}. A challenge in DLRMs is how to deal with sparse categorical features. For instance, a single categorical feature in the YouTube recommendation model contains tens of millions of video IDs~\cite{covington2016deep}.  To handle the categorical features, modern DLRMs use embedding tables, which are hash tables that map a categorical index to a vector.

Unfortunately, embedding tables are often the storage and efficiency bottlenecks in production-scale DLRMs. On the one hand, the embedding tables can be extremely large. For example, the embedding tables in the Meta DLRMs demand multi-terabyte memory~\cite{acun2021understanding,mudigere2022software}. Thus, modern distributed training systems for DLRMs often have to adopt model parallelism to partition the tables and place them on different devices, such as GPUs and CPUs~\cite{acun2021understanding,covington2016deep,liu2017related,gomez2015netflix,lian2022persia}. On the other hand, embedding tables often incur significant computation and communication costs. For instance, it is reported that embedding tables account for 48\% of the total computation and 65\% of the total communication costs in one of the Meta DLRMs~\cite{zha2022dreamshard}.

\begin{figure*}[t]
    \centering
    \includegraphics[width=0.95\textwidth]{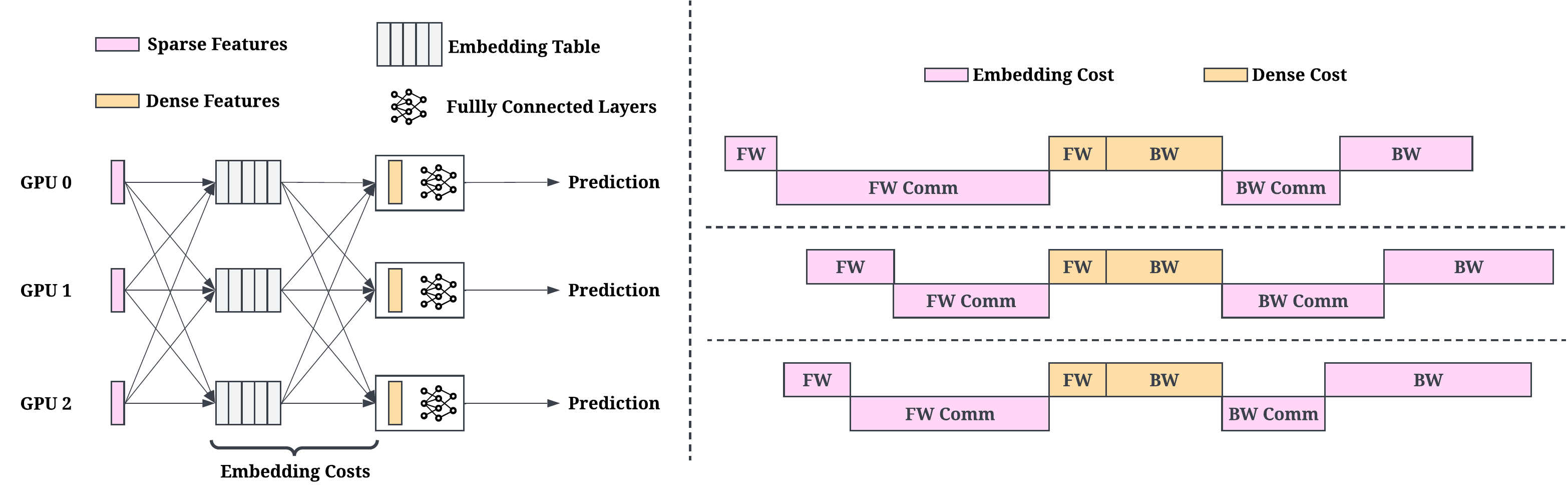}
    \vspace{-5pt}
    \caption{Left: an illustrative distributed training workflow of DLRMs~\cite{naumov2019deep} on three GPUs. Right: typical GPU traces in the fully synchronous mode~\cite{mudigere2022software}. For each GPU, the top and bottom threads are for computation and communication, respectively, and they can overlap. FW and BW stand for forward and backward computations. FW Comm and BW Comm mean forward and backward communications. We only visualize the major costs; the other costs are often neglectable or hidden due to overlapping. The embedding forward operations (pink FW) often do not start at the same time across GPUs because of the different ending times of the embedding backward operations (pink BW) in the previous training iteration.}
    \label{fig:illustration}
\end{figure*}

The left side of Figure~\ref{fig:illustration} shows the computation and communication costs for embedding tables in a typical distributed training workflow of DLRMs\footnote{This work focuses on sharding among GPU devices. We will study CPU or mixed CPU-GPU sharding scenarios in the future.}~\cite{naumov2019deep}. It exploits a combination of model parallelism, i.e., partitioning embedding tables and placing them to multiple GPUs, and data parallelism, i.e., duplicating the fully connected layers and partitioning the training data. In the forward pass, each GPU queries the other GPUs with its sparse features to look up the embeddings from their tables (forward computation) and obtain the embeddings through an all-to-all communication (forward communication). In the backward pass, the gradients are sent back to the GPUs with another all-to-all communication (backward communication) and applied to the embeddings (backward computation).

An important design factor that can significantly impact the embedding costs is embedding table sharding, i.e., the strategy of partitioning and placing embedding tables. If not carefully partitioned, the embedding tables can easily lead to imbalances, where some devices have significantly more computation and communication costs than others, leading to a straggler effect in the synchronous training setting~\cite{mudigere2022software}. While some heuristic-based sharding strategies have been proposed~\cite{acun2021understanding,lui2021understanding}, they rely on oversimplified cost functions so they often have unsatisfactory sharding performance.


Recently, reinforcement learning (RL) has shown promise in embedding table sharding~\cite{zha2022autoshard,zha2022dreamshard}. The idea is to make sharding an optimization problem, which aims to identify a sharding plan that can minimize the overall cost. These methods formulate sharding as a Markov decision process (MDP), whose states and rewards are computation and communication costs estimated by neural networks. Then they train another policy network to solve the MDP to minimize the overall embedding costs. These learning-based methods have achieved a significant improvement over the heuristic sharding strategies~\cite{zha2022autoshard,zha2022dreamshard}.

Despite the successes of RL-based approaches, it is difficult to deploy them. \textbf{1)} They only consider table-wise sharding, i.e., they treat tables as the smallest units in sharding and focus on how to assign each table to a device. However, it is very likely that one table is extremely large or costly, which makes it a memory and computation bottleneck. Adopting these approaches may lead to an out-of-memory error or an undesirable balance. \textbf{2)} The policy network is often trained on very few sharding tasks, so frequent re-training will be needed to handle unseen tasks. \textbf{3)} The policy in RL is notoriously unstable with a high variance~\cite{henderson2018deep,zha2019experience,zha2021douzero,lai2020dual}; that is, even if we train the same policy on the same MDP with multiple independent runs, some of the runs may work well but the others could fail. However, we often demand a stable sharding solution in production.

\begin{figure}[t]
    \centering
    \includegraphics[width=0.48\textwidth]{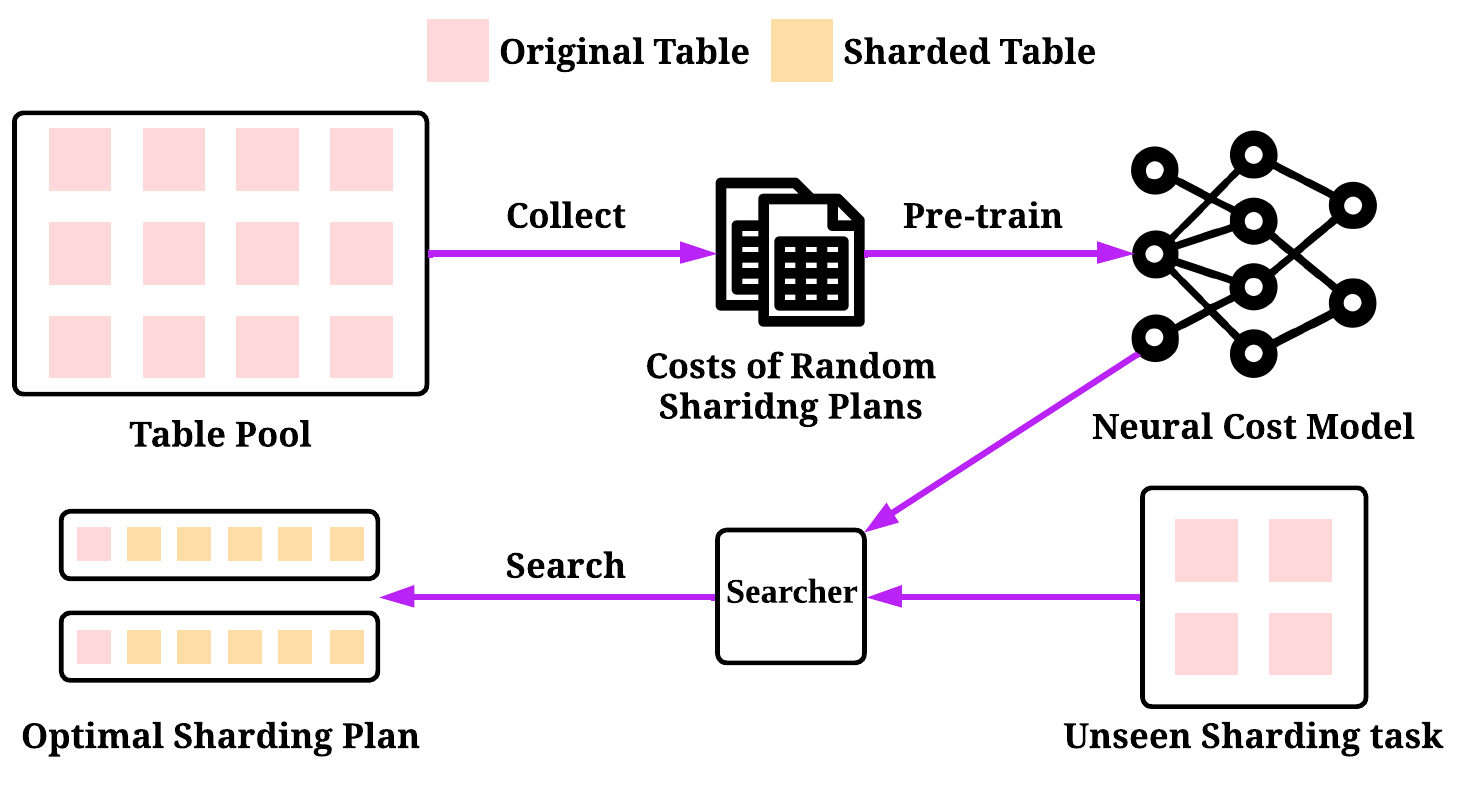}
    \caption{NeuroShard identifies the optimal sharding plan with \emph{``pre-train, and search''}.}
    \label{fig:concept}
\end{figure}

Motivated by the recent successes of ``pre-train, prompt, and predict'' in large language models~\cite{liu2023pre,brown2020language,touvron2023llama,chuang2023spec,tang2023science}, we explore a \emph{``pre-train, and search''} paradigm for efficient sharding and present NeuroShard, illustrated in Figure~\ref{fig:concept}. To handle the extremely large or costly tables, we incorporate \emph{column-wise} sharding into the optimization process, where a table can be partitioned into two smaller tables, each with half the columns of the original one. Unlike RL-based methods that stochastically train a policy network, we pre-train general neural cost models on augmented data to cover comprehensive sharding scenarios. Once trained, the cost models serve as a simulator to efficiently estimate the embedding costs for any sharding plans. Built upon the pre-trained cost models, NeuroShard identifies the best column-wise and table-wise sharding plans with beam search and greedy grid search, respectively. NeuroShard not only outperforms the state-of-the-art but also can be easily deployed in production. In summary, we make the following contributions.
\begin{itemize}
    \item We provide a comprehensive analysis of the computation and communication costs of embedding tables. We observe: \textbf{1)} partitioning a table column-wisely will increase the overall cost so that column-sharding has a tradeoff between overall cost and balance, \textbf{2)} multi-table computation cost has a non-linear correlation with the sum of single-table costs, and \textbf{3)} the communication cost is mainly determined by table dimensions.
    \item Motivated by the observations, we devise NeuroShard, a \emph{``pre-train, and search''} framework which searches for column-wise and table-wise sharding plans on pre-trained cost models. Unlike the RL-based methods whose cost models only have limited coverage, our cost models are trained for a once-for-all purpose, i.e., training one universal cost model for all the sharding tasks. Moreover, our cost models can readily support column-wise sharding since it is trained on an augmented table pool with various table dimensions. 
    \item NeuroShard achieves up to 23.8\% improvement over the state-of-the-art without out-of-memory error on the benchmark sharding dataset~\cite{naumov2019deep}\footnote{\label{foot:dlrm}\url{https://github.com/facebookresearch/dlrm_datasets}}.
    \item NeuroShard has been deployed to an ultra-large production DLRM to shard multi-terabyte embedding tables to hundreds of GPUs. NeuroShard achieves an 11.6\% improvement in embedding costs, which translates to 6.6\% end-to-end training throughput improvement, which is a significant  speedup since the production DLRM has been heavily optimized.
\end{itemize}

\section{Understanding Embedding Cost}
\label{sec:2}

This section analyzes the embedding costs to motivate the sharding algorithm design. The right-hand side of Figure~\ref{fig:illustration} shows a typical trace in one training iteration, which mainly consists of the computation/communication of the embeddings and the computation of the fully connected layers. Note that there are some other costs that are not visualized in the figure, such as the communication of the sparse features before embedding lookup, and the weight synchronization of the fully connected layers, etc. These costs are often neglectable or hidden because of the overlapping of computation and communication.


We first explain why imbalances can cause more embedding costs using the example trace. In the trace of GPU 1, the embedding backward communication and computation take more time than the other GPUs, which makes its embedding forward computation in the next iteration start later than its peer GPUs. The embedding forward computation of GPU 1 again takes a longer time than those of the other GPUs. As such, the above delays are finally accumulated, forcing GPU 1 to start embedding forward communication significantly later than other GPUs. This imbalance issue results in significant idle times for GPU 0 and GPU 2. 

To reduce the accumulated delay, we need to balance the computation and communication costs associated with embeddings across all GPUs. In the following, we analyze the costs separately. All the results are collected on 2080Ti GPUs with a modern embedding implementation from FBGEMM~\cite{fbgemm}, which fuses multiple table lookups as a single operation. Appendix~\ref{appendix:A} provides more details on all the analytical experiments.

\begin{figure}[t]
  \centering
  \begin{subfigure}[b]{0.24\textwidth}
    \centering
    \includegraphics[width=0.95\textwidth]{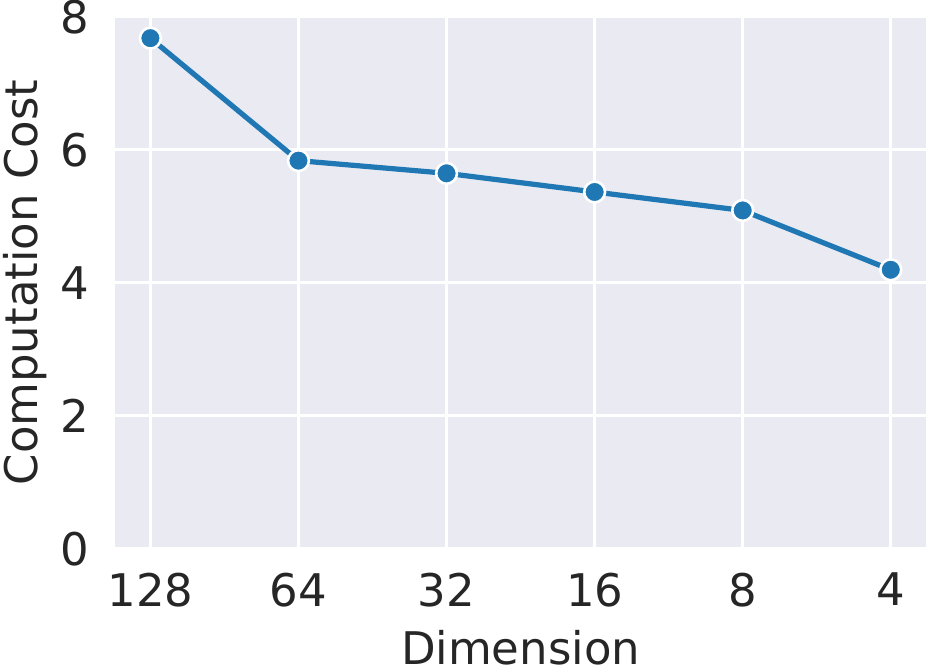}
  \end{subfigure}%
  \begin{subfigure}[b]{0.24\textwidth}
    \centering
    \includegraphics[width=0.95\textwidth]{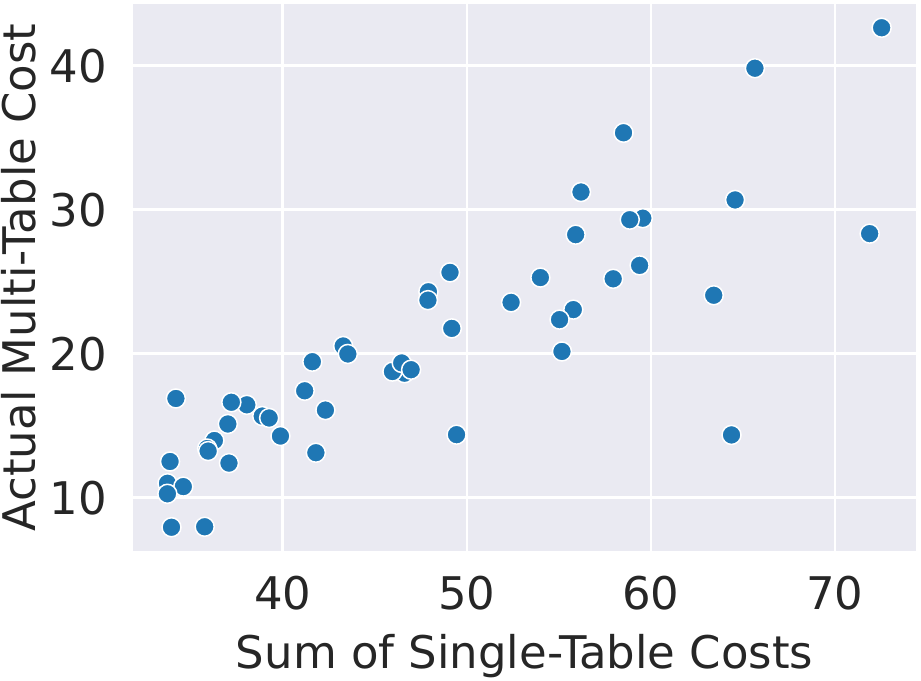}
  \end{subfigure}%
  \caption{Left: computation costs w.r.t. dimensions. We observe similar patterns for other tables (see Appendix~\ref{appendix:A1}). Right: the sum of single-table costs versus the actual multi-table cost.}
  \label{fig:computeanalysis}
\end{figure}

\subsection{Computation Cost Analysis}

Computation costs are mainly decided by the table configurations and the lookup indices. Some important factors were used to quantify the computation costs~\cite{zha2022autoshard}. \textbf{1) Dimension:} the number of columns of the table. A larger dimension often means a higher computation cost since it technically leads to more memory bandwidth use. \textbf{2) Hash size:} the number of rows of the table. It indirectly impacts the computation costs by affecting the caching/prefetching behaviors. \textbf{3) Pooling factor:} the number of embedding indices in a lookup. We often calculate the mean pooling factor of a batch of indices. A higher mean pooling factor leads to a higher computation cost since it determines the workloads of the lookup. \textbf{4) Indices distribution:} the access pattern and distribution of the lookup, i.e., some indices can be accessed more frequently than others. It indirectly impacts the costs by affecting cache effectiveness. Also, the number of unique embeddings accessed in a batch can also influence the caching. Fewer indices being accessed will often lead to smaller costs.

The previous analysis suggests that the actual computation costs have complex and non-linear correlations with the above factors~\cite{zha2022autoshard}, which makes cost estimation hard. Here, we perform two analytical experiments to provide a deeper understanding and motivate the algorithm designs of column-wise sharding and table-wise sharding.

First, we study the impact of dimension. We randomly choose a table from the benchmark sharding dataset named DLRM~\cite{naumov2019deep}$^{\ref{foot:dlrm}}$. The left-hand side of Figure~\ref{fig:computeanalysis} visualizes its computation costs (forward + backward) with varying dimensions of $\{128, 64, 32, 16, 8, 4\}$. As expected, a larger dimension corresponds to a higher computation cost. However, we also have the following observation.

\begin{theorem}
\label{obs:1}
When partitioning a table into two halves column-wisely, the computation cost of each shard is larger than half the cost of the original table.
\end{theorem}

For example, the cost of dimension 64 is much larger than the half of cost of dimension 128. This could be explained by parallelism and operation fusion; the fused embedding table operation can achieve better optimization in the CUDA kernel than in each operation alone. This also indicates a trade-off in column-wise sharding: while partitioning tables into smaller tables could improve load balance, it may increase the overall computation cost. Thus, the column-wise sharding algorithm needs to strike a balance between the load balance and the overall computation cost.


Then, we investigate whether we can estimate the computation cost of a multi-table operation with the sum of the single-table costs. This is important because if this assumption holds, we could well balance the computation costs with tools such as mixed integer linear program~\cite{sethi2022recshard}. We randomly sample 50 subsets of tables from the DLRM dataset, where each subset contains 10 tables. Then we plot their relationships on the right-hand side of Figure~\ref{fig:computeanalysis}.

\begin{theorem}
\label{obs:2}
Multi-table computation cost has a non-linear relationship with the sum of single-table costs.
\end{theorem}

Evidently, table-wise sharding algorithms must consider the nonlinearity of the multi-table costs.

\begin{figure}[t]
  \centering
  \begin{subfigure}[b]{0.24\textwidth}
    \centering
    \includegraphics[width=0.95\textwidth]{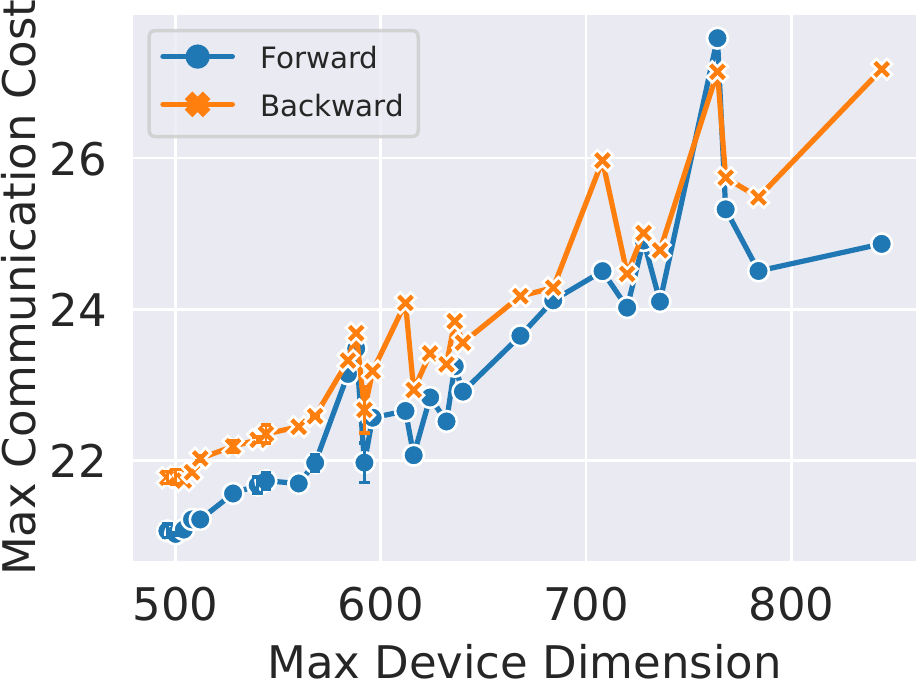}
  \end{subfigure}%
  \begin{subfigure}[b]{0.24\textwidth}
    \centering
    \includegraphics[width=0.95\textwidth]{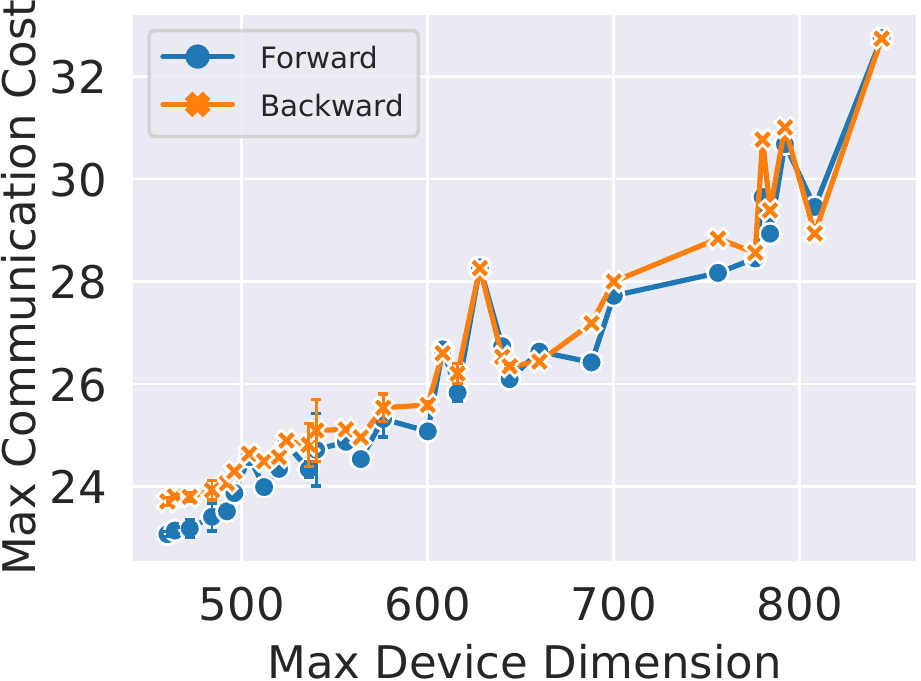}
  \end{subfigure}%
  \caption{Max forward/backward communication cost versus max device dimensions among 4 GPUs (left) and 8 GPUs (right).}
  \label{fig:commanalysis}
\end{figure}

\begin{figure}[t]
    \centering
    \includegraphics[width=0.48\textwidth]{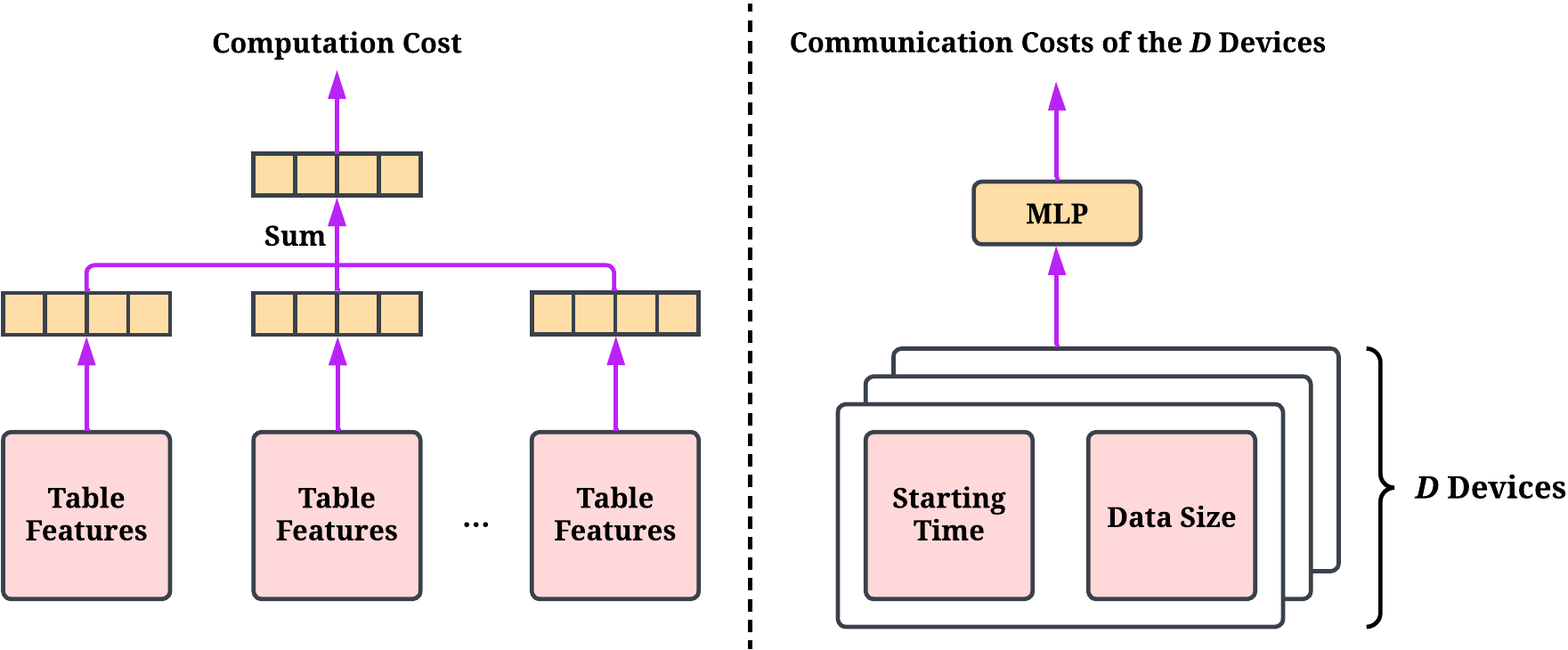}
    \caption{The neural architectures of the computation cost model (left) and the communication cost model (right).}
    \label{fig:costmodels}
\end{figure}

\subsection{Communication Cost Analysis}

In distributed training, every pair of GPUs need to communicate in both the forward and backward passes. Thus, each GPU could have a different communication latency. Our goal is to minimize the max communication cost among all the GPUs since the slowest one will become the bottleneck. Intuitively, the communication costs depend on the sizes of the data to be sent in the GPUs, where the data size of a GPU can be estimated by the product of the batch size and the device dimension, which is defined as the sum of the dimensions of the tables in the device. Since all the GPUs have the same batch size, the device dimension becomes the determining factor of communication balance.


We conduct an experiment to understand the relationship between the max device dimension and the max forward/backward communication cost. We randomly sample a subset of tables (40 tables for 4 GPUs, and 80 tables for 8 GPUs) from the DLRM dataset and select a random dimension for each table from $\{128, 64, 32, 16, 8, 4\}$. Then we shard these tables to 4 or 8 GPUs with varying max device dimensions. We benchmark 50 assignments in Figure~\ref{fig:commanalysis}.

\begin{theorem}
\label{obs:3}
The max forward/backward communication cost among all the GPUs positively correlates with the max device dimension among all the GPUs.
\end{theorem}

The above observation motivates us to adopt an alternative way to balance communication costs, i.e., minimizing the max device dimension among all the GPUs. Balancing the device dimensions is much easier since the device dimension is simply the sum of the dimensions of the tables.

\begin{figure*}[t]
    \centering
    \includegraphics[width=0.96\textwidth]{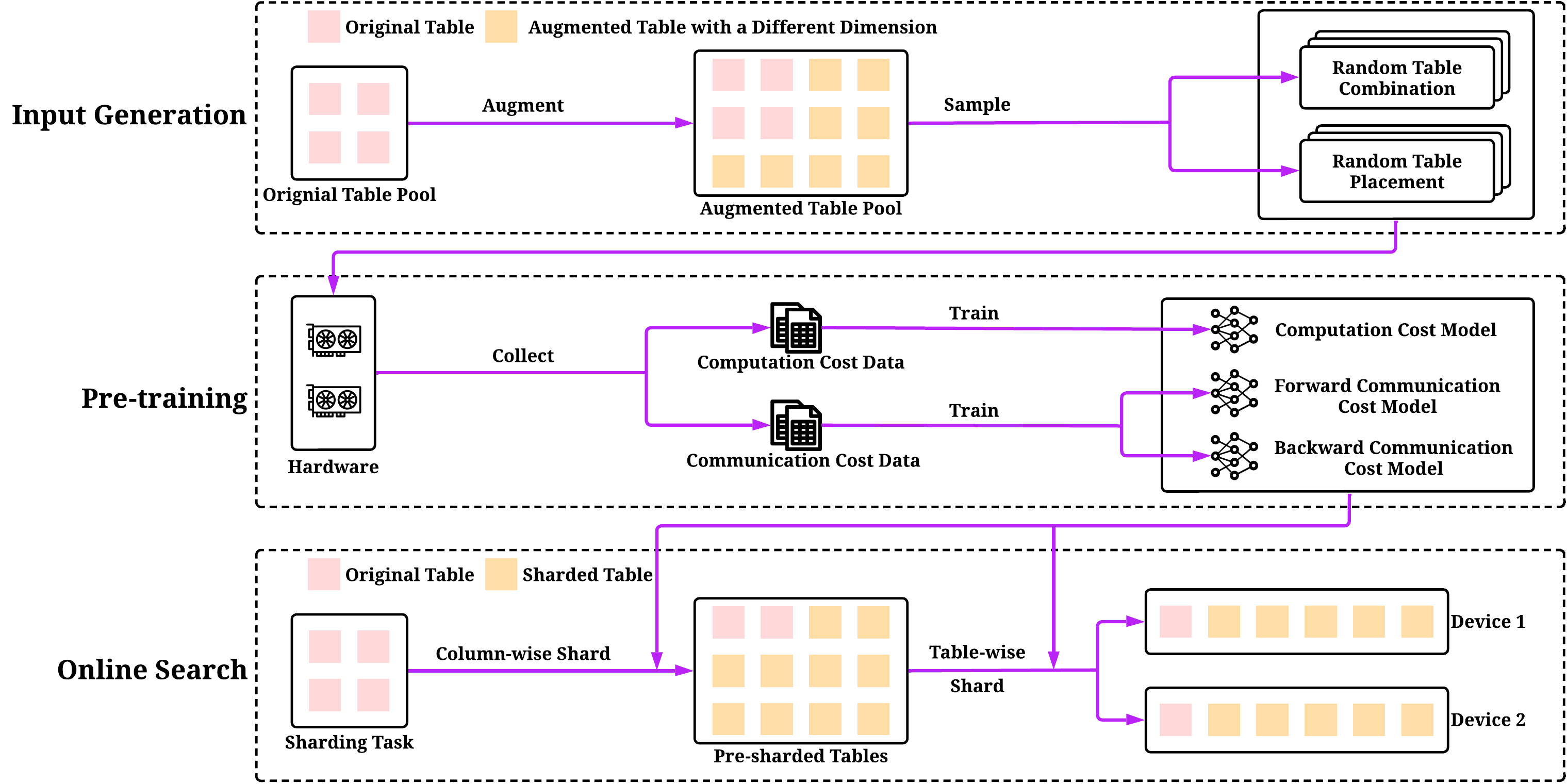}
    \vspace{-8pt}
    \caption{Overall workflow of NeuroShard. We first generate random inputs based on the augmented tables (top row). Then we run a micro-benchmark to collect the costs and pre-train three neural networks to predict the costs (middle row). Finally, we perform an online search based on the pre-trained cost models for embedding table sharding (bottom row). }
    \vspace{-2pt}
    \label{fig:overview}
\end{figure*}

\section{NeuroShard Framework}
\label{sec:3}

Motivated by the three observations above, we propose NeuroShard, an embedding table sharding framework based on pre-trained neural cost models and online search. Figure~\ref{fig:overview} shows the workflow. The main idea is to pre-train neural networks to predict the computation and communication costs, which can serve as a sharding simulator to quickly estimate the embedding costs for any sharding tasks and any sharding plans. Then we perform an online search on the pre-trained cost models to identify the best sharding plan without real GPU execution. In what follows, we describe how to generate synthetic inputs for training data collection~(Section~\ref{sec:31}) and how to train neural networks for cost prediction~(Section~\ref{sec:32}). Once the cost models are pre-trained, we then present how to perform an online search to optimize column-wise sharding and table-wise sharding~(Section~\ref{sec:33}).

\subsection{Generating Synthetic Inputs}
\label{sec:31}
Pre-training is mainly a data-centric procedure~\cite{zha2023data-centric-survey,zha2023data-centric-perspectives}, where high-quality cost data plays an essential role. This subsection describes how to generate table inputs that can cover different table combinations and placements for benchmarking the computation and communication costs, which includes table augmentation, random table combination generation, and random table placement generation. Note that to achieve the best coverage, the generation strategy should consider the infrastructure for model training and the embedding tables in the model. For example, if a model has a lot of embedding tables but only a very limited number of GPUs is used, the generated inputs should cover the table combinations that have lots of tables. In the following, we mainly discuss the high-level strategy on how to achieve good coverage. We will introduce the instantiation of the strategy on the DLRM dataset in Section~\ref{sec:4}.

\textbf{Table augmentation.} In DLRMs, each embedding table corresponds to a sparse feature, which is often collected from users or items. In real-world applications, we often have a pool of embedding tables, where the machine learning engineers will perform feature selection, i.e., choosing a subset of embedding tables from the pool for model training. In this procedure, the dimensions of the tables could be adjusted. Further, column-wise sharding will also generate new tables with different dimensions. Thus, the generated table inputs should be able to cover tables with different dimensions. To achieve this, we perform table augmentation. Specifically, for each table, we generate augmented tables with different dimensions. For example, suppose the original table has a dimension of 64, we could generate 5 augmented tables with dimensions of 128, 32, 16, 8, and 4 to accommodate the potential dimension adjustment in the model design and column-wise sharding. The table augmentation results in an augmented table pool, which will be used for data generation. Appendix~\ref{appendix:B1} summarizes the detailed augmentation process.


\textbf{Random table combination generation.} The table combinations will be used to benchmark computation costs. The generated combinations should cover different numbers of tables on a GPU. To achieve this, we first uniformly sample the number of tables $T$ in a certain range and then randomly select a subset of $T$ tables from the augmented table pool. Appendix~\ref{appendix:B2} summarizes the detailed generation process.

\textbf{Random table placement generation.} The table placements will be used to benchmark communication costs. The generated placements should cover different degrees of balance for the device dimensions and different communication starting timestamps. \textbf{1) To simulate different degrees of balance, we adopt a greedy strategy equipped with randomness.} Specifically, we first randomly sample a subset of tables from the table pool and sort the tables in descending order based on the dimension. Starting from the table with the largest dimension, with a probability of $p$, where $p$ is uniformly sampled in $[0, 1]$ for each table placement, we assign the current table to the GPU with the lowest sum of the table dimensions so far, and with a probability of $(1-p)$, we randomly assign the current table. Here, $p$ can indirectly control the degree of balance. When $p=1$, this strategy will greedily balance dimensions in each step so the sharding plan can well balance the dimensions. When $p=0$, the sharding plan will be random so that the dimensions will be very likely to be imbalanced. Since $p$ is randomly selected, we can cover sharding plans with different degrees of balance. \textbf{2) We randomly generate the communication starting timestamp of each GPU.} Recall that in the trace analysis (the right-hand side of Figure~\ref{fig:illustration}), the delays caused by previous operations can be accumulated to make the forward communication start significantly later than the other GPUs. Naturally, we also need to take the delays into account when benchmarking the communication costs since the communication could have different behaviors when they do not start simultaneously across the GPUs. To simulate the delays, we randomly select a starting timestamp in a certain range for each GPU. Appendix~\ref{appendix:B3} summarizes the detailed generation procedure.


\subsection{Pre-training Neural Cost Models}
\label{sec:32}
Given the generated inputs, we run micro-benchmark, e.g., PARAM Benchmarks\footnote{\url{https://github.com/facebookresearch/param}}, to collect the actual computation and communication costs for cost model training. Appendix~\ref{appendix:C} provides more details of cost model training.

Figure~\ref{fig:costmodels} illustrates the neural architectures of the computation and communication cost models. \textbf{1) Computation cost model:} the architecture follows~\cite{zha2022autoshard}. Specifically, each table is represented with some features, including dimension, hash size, pooling factor, and indices distribution. Given a table combination, we use a shared MLP to process all the table features to obtain table representations. We obtain a fixed-dimension representation of a table combination by performing an element-wise sum of all the table representations. Finally, we use another MLP to produce the computation costs (forward + backward). \textbf{2) Communication cost model:} we use an MLP to predict the communication costs of all the GPUs based on the starting timestamps and the transferred data sizes. We train two separate models for forward and backward communications.


\textbf{Deployment of the neural cost models.} In real-world DLRMs, the indices distributions could shift over time. Thus, we may need to re-train/fine-tune and redeploy the cost models to tackle the potential shifts. To enable a smooth re-deployment, we often need to have strict version control to ensure that one training job is always associated with the same version of cost models. This is particularly important for checkpointing since we need a consistent sharding plan when resuming the training. We find a re-training interval of three months is sufficient in our production environment. One could also periodically calculate the prediction errors of the cost model by sampling a batch of table indices and trigger re-training or fine-tuning when the error exceeds a certain threshold. Note that we often only need to re-train when the indices distributions shift. Re-training is not needed when table dimension changes, as the table augmentation has already encompassed various table dimensions.

\subsection{Online Search}
\label{sec:33}
\begin{figure}[t]
    \centering
    \includegraphics[width=0.4\textwidth]{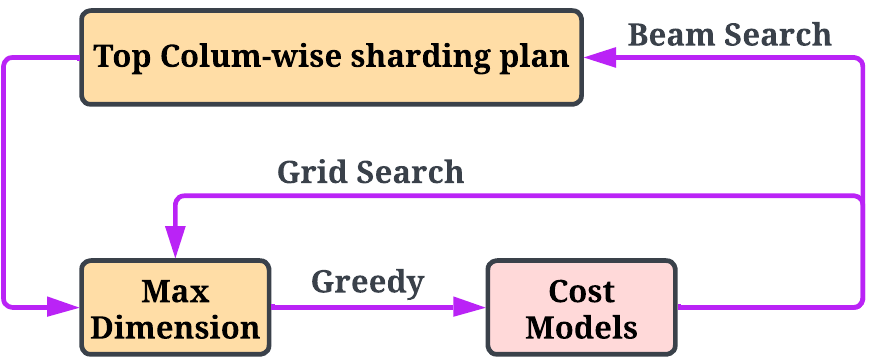}
    \vspace{-5pt}
    \caption{The search process. The outer loop finds the best column-wise sharding plan with beam search. The inner loop focuses on the max dimension constraint of greedy allocation with grid search.}
    \vspace{-5pt}
    \label{fig:search}
\end{figure}

The pre-trained cost models serve as a universal simulator for embedding table sharding. They can estimate the embedding cost of any sharding plan for any sharding task efficiently by summing up the predicted computation, forward communication, and backward communication costs. In this subsection, we introduce how to leverage this simulator to minimize the embedding cost with search. Figure~\ref{fig:search} shows an overview. In the outer loop, we search the column-wise sharding plan with beam search. In the inner loop, we find the best max dimension constraint with a greedy grid search for table-wise sharding. The presented search algorithm is mainly motivated by the observations in Section~\ref{sec:2}.

\textbf{Optimization problem formulation.} We denote the column-wise sharding plan as $\mathbf{c}=[c_1, c_2, ..., c_m]$, where $c_i$ means, in step $i$, we shard the table of index $c_i$ into two halves column-wisely and append the resultant new table to the end of the table list. Let $\mathbf{t}=[t_1, t_2, ..., t_T]$ denote the table-wise sharding plan that assigns $T$ tables to $D$ GPU devices, where $t_i \in \{1, 2, ..., D\}$. $\mathbf{t}$ depends on $\mathbf{c}$ since $\mathbf{t}$ operates on the column-wise sharded tables. Both $\mathbf{c}$ and $\mathbf{t}$ must satisfy some constraints. For example, the embedding operations in FBGEMM~\cite{fbgemm} require that the dimension must be dividable by 4. For $\mathbf{c}$, the sharding plan has to satisfy the GPU memory constraints. We denote their legal plan spaces as $\mathcal{C}$ and $\mathcal{T}$, respectively, where $\mathcal{T}$ depends on the selected $\mathbf{c}$. The objective is
\begin{equation}
    \argmin_{\mathbf{c} \in \mathcal{C}, \mathbf{t} \in \mathcal{T}} f(\mathbf{c}, \mathbf{t}).
\end{equation}
$f(\mathbf{c}, \mathbf{t})$ is the simulated embedding cost. The dependency of $\mathbf{t}$ on $\mathbf{c}$ naturally makes the search of $\mathbf{c}$ as the outer loop and the search of $\mathbf{t}$ as the inner loop.

\textbf{Column-wise sharding with beam search.} Column-wise sharding can remove oversized tables and costly tables to enable a better balance. However, from Observation~\ref{obs:1}, column-wise sharding increases the overall computation cost. Thus, the desirable sharding plan should enable a balance with minimum steps. We propose a beam search strategy to reduce the search space. The main idea is that we often only need to column-wisely shard tables with large sizes and high computation costs. Specifically, in each iteration $i$, we identify the top $N$ costly tables and the top $N$ tables with the largest sizes as the candidates (with duplicates removed). Similarly, we only consider the top $K$ best column-wise sharding plans found in the previous step. For each of the top $K$ plans, we add each of the candidate tables to it to obtain a new sharding plan and run the inner loop to get the cost. We again identify the top $K$ best new sharding plans for the next sharding step. We perform $L$ sharding steps and output the sharding plan with the minimum cost. $L$, $K$, and $N$ are hyperparameters to balance optimality and efficiency.


\begin{algorithm}[t]
\caption{BeamSearch}
\label{alg:1}
\setlength{\intextsep}{0pt} 
\begin{algorithmic}[1]
\STATE \textbf{Input:} $T$ embedding tables, beam search hyerperparameters ($N$, $K$, and $L$)
\STATE Best global column-wise sharding plan $\mathbf{c}^*$ $\leftarrow []$
\STATE Best global table-wise sharding plan $\mathbf{t}^*$ $\leftarrow NULL$
\STATE Best column-wise sharding plans $\mathcal{C}_p$ $\leftarrow \{[]\}$
\STATE Initialize a global cache $global\_cache$

\FOR{$outer\_loop$ = 1, 2, ..., $L$}
    \STATE Column-wise sharding plans in the next step $\mathcal{C}'_p \leftarrow []$
    \FOR{each plan $\mathbf{c}_p$ in $\mathcal{C}_p$}
        \STATE Obtain the candidate tables based on $\mathbf{c}_p$ by merging the top $N$ costly tables and the top $N$ tables with the largest sizes with duplicates removed
        \FOR{each candidate table $t$}
            \STATE $col\_plan$ $\leftarrow$ $\mathbf{c}_p$ with $t$ appended in the end
            \STATE $cost$, $\mathbf{t}$ $\leftarrow$ GreedyGridSearch($global\_cache$, $col\_plan$)
            \STATE Append ($col\_plan$, $cost$) to $\mathcal{C}'_p$
            \IF{a lower $cost$ is observed}
                \STATE $\mathbf{c}^*$ $\leftarrow col\_plan$
                \STATE $\mathbf{t}^*$ $\leftarrow \mathbf{t}$
            \ENDIF
        \ENDFOR
    \ENDFOR
    \STATE $\mathcal{C}_p \leftarrow$ plans with the top $K$ lowest cost in $\mathcal{C}'_p$
\ENDFOR
\STATE Return $\mathbf{c}^*$, $\mathbf{t}^*$

\end{algorithmic}
\end{algorithm}

\textbf{Table-wise sharding with greedy grid search.} Given a list of column-wisely sharded tables, we describe how to perform table-wise sharding in the inner loop. We propose a grid search strategy to find the best balance of computation and communication costs based on two ideas. \textbf{1)} Motivated by Observation~\ref{obs:2}, we propose a greedy algorithm to balance the multi-table computation costs. \textbf{2)} Inspired by Observation~\ref{obs:3}, we make the max device dimension a constraint for the greedy algorithm to achieve the communication balance, where the best max device dimension is identified with grid search. Specifically, in each step we execute the following: \textbf{1)} Choose a max device dimension $max\_dim$. \textbf{2)} Sort the tables in descending order based on the computation cost predicted by the cost model. \textbf{3)} Starting from the table with the highest cost, we assign tables one by one to the device with the lowest device cost so far subject to the memory and $max\_dim$ constraints, where the device cost is predicted by the cost model. \textbf{4)} Evaluate the embedding cost with the cost models. We grid search $max\_dim$ as follows. Given a starting value $M_s$, an ending value $M_e$, and the total number steps $M$, we try all the values in $[M_s, M_e]$ with a step size of $(M_e - M_s) / (M-1)$. We empirically fix $M_s$ to be the average dimension across device and $M_e$ to be $1.5 * M_s$. $M$ is a hyperparameter to control the granularity of the search.

\textbf{Implementation with caching.} The most expensive part in the search is predicting the computation cost. It needs to be called for $O(LKNMTD)$ times ($T$ is the number of tables, and $D$ is the number of GPUs), where each call requires a forward pass of the cost model. Fortunately, we find there are lots of duplicated calls in the search. This is because if we only make small changes to the column-wise sharding plan or $max\_dim$, the cost model will be very likely to be asked to predict the cost for the same set of tables in most of the steps. Thus, we can naturally use a life-long hash map as a cache for acceleration. In practice, the cache hit rate can reach 95\% (see Table~\ref{tab:ablation}). We summarize the beam search in Algorithm~\ref{alg:1} and greedy grid search in Algorithm~\ref{alg:2}.

\begin{algorithm}[t]
\caption{GreedyGridSearch}
\label{alg:2}
\setlength{\intextsep}{0pt} 
\begin{algorithmic}[1]
\STATE \textbf{Input:} $T$ embedding tables, $D$ GPU devices, pre-trained neural cost models, grid search hyperparameter $M$, $global\_cache$, $col\_plan$
\STATE Generate $T'$ column-wise sharded tables using $col\_plan$ where $T' = T + |col\_plan|$
\STATE Sort the $T'$ tables in descending order based on the costs predicted by the computation cost model
\STATE Best table-wise sharding plan $\mathbf{t}^*$ $\leftarrow NULL$
\STATE Best computation cost $cost^* \leftarrow Inf$
\FOR{$inner\_loop$ = 1, 2, ..., $M$}
   \STATE Get $max\_dim$ based on $M$
   \STATE Initialize table-wise sharding plan $\mathbf{t} \leftarrow []$
   \FOR{each of the $T'$ tables}
        \STATE Get candidate GPUs that will not cause memory error with device dimension smaller than $max\_dim$
        \FOR{each of the candidate GPUs}
            \IF{the tables in GPU are in $global\_cache$}
                \STATE Get $cost$ from $global\_cache$
            \ELSE
                \STATE Get $cost$ with the computation cost model and store the cost into $global\_cache$
            \ENDIF
            \STATE Append the GPU with the lowest cost to $\mathbf{t}$
            \IF{a lower $\mathbf{t}$ is observed}
                \STATE $\mathbf{t}^*$ $\leftarrow \mathbf{t}$, $cost^* \leftarrow cost$
            \ENDIF
        \ENDFOR
   \ENDFOR
\ENDFOR
\STATE Return $cost^*$, $\mathbf{t}^*$

\end{algorithmic}
\end{algorithm}

\begin{table*}[t]
    \centering
    \footnotesize
    \caption{Embedding table cost in milliseconds (averaged over 100 randomly constructed sharding tasks) of NeuroShard against baselines with 4 or 8 GPUs and maximum table dimensions from 4 to 128. The top-1 and top-2 results are highlighted in boldface and underlined, respectively. ``-" indicates that the method cannot scale, i.e., at least one of the 100 tasks suffers from memory explosion. The bottom row summarizes the improvement of NeuroShard over the strongest baseline.}
    \label{tab:performance}
    \setlength{\tabcolsep}{1.7pt}
    \begin{tabular}{l|l|cccccc|cccccc}
    \toprule
     
    \multirow{2}{*}{Category} & \multirow{2}{*}{Method} & \multicolumn{6}{c|}{4 GPUs} & \multicolumn{6}{c}{8 GPUs} \\
    \cline{3-14}
    ~ & ~ & 4 & 8 & 16 & 32 & 64 & 128 & 4 & 8 & 16 & 32 & 64 & 128 \\
    \midrule
    \midrule

     Random & - & 25.47 & 30.10 & - & - & - & - & 33.40 & 36.43 & - & - & - & - \\
     \midrule
     \multirow{4}{*}{Greedy} & Size-based & 24.45 & 29.60 & 30.86 & 37.80 & 41.59 & - & 31.75 & 34.30 & 37.70 & 46.07 & 54.57 & - \\
     ~ & Dim-based & 23.51 & 28.46 & 29.76 & 35.98 & 38.71 & -& 27.54 & 32.20 & 34.78 & 42.35 & 47.54 & - \\
     ~ & Lookup-based & 18.69 & 24.34 & 26.83 & 34.62 & 38.69 & - & 21.83 & 26.66 & 30.59 & \underline{39.07} & 47.47 & - \\
     ~ & Size-lookup-based & 18.38 & 24.18 & 26.81 & 33.94 & - & - & 21.27 & 26.55 & 30.34 & 39.28 & 48.17 & - \\
     \midrule
     Reinforcement & AutoShard & \underline{17.99} & \underline{22.08} & - & - & - & - & \underline{20.79} & - & - & - & - & - \\
     Learning & DreamShard & 18.78 & 22.59 & \underline{25.08} & \underline{30.74} & - & - & 21.40 & \underline{25.30} & \underline{26.90} & - & - & - \\
     \midrule
     Planning & TorchRec & 19.20 & 26.00 & 28.24 & 34.88 & \underline{38.13} & \underline{47.22} & 22.34 & 28.99 & 32.71& 39.90 & \underline{47.43} & 60.58 \\
     \midrule
     Cost Modeling & NeuroShard & \textbf{17.74} & \textbf{21.75} & \textbf{23.11} & \textbf{28.86} & \textbf{31.55} & \textbf{39.99} & \textbf{20.68} & \textbf{23.23} & \textbf{25.64} & \textbf{32.30} & \textbf{38.30} & \textbf{49.10} \\
     \midrule
    \midrule
    \multicolumn{2}{c|}{Improvement of NeuroShard} & +1.4\% & +1.5\% & +8.5\% & +6.5\% & +20.9\% & +18.1\% & +0.5\% & +8.9\% & +4.9\% & +21.0\% & +23.8\% & +23.4\% \\
   
     \bottomrule
    \end{tabular}
\end{table*}

\section{Experiments}
\label{sec:4}

The experiments aim to answer the following research questions. \textbf{RQ1:} How does NeuroShard compare with the state-of-the-art sharding algorithms~(Section~\ref{sec:41})? \textbf{RQ2:} How accurate are the neural cost models~(Section~\ref{sec:42})? \textbf{RQ3:} How does each search design contribute to the performance~(Section~\ref{sec:43})? \textbf{RQ4:} How do the hyperparameters impact the performance~(Section~\ref{sec:44})? \textbf{RQ5:} Can NeuroShard boost end-to-end training throughput~(Section~\ref{sec:45})?

\textbf{Datasets.} The public large datasets (e.g., Criteo, Avazu, and KDD) often do not match the industrial-scale data and are unsuitable for evaluating sharding algorithms. Please find more discussions in Appendix~\ref{appendix:D}. Following the previous work~\cite{zha2022autoshard,zha2022dreamshard}, we use the DLRM dataset~~\cite{naumov2019deep}$^{\ref{foot:dlrm}}$, which contains 856 synthetic tables whose indices distributions are similar to the production workloads in Meta. These 856 tables serve as the table pool. \textbf{We construct synthetic sharding tasks as diversely as possible to test sharding algorithms in different scenarios.} We consider two sets of sharding tasks, which aim to shard tables to 4 and 8 GPUs, and each GPU has a memory constraint of 4 GBs for embedding tables. We randomly sample a dimension for each table in $\{4, 8, ..., 2^j\}$, where $2 \le j \le 7$, and $2^j$ specifies the maximum possible dimension for a table. A larger $2^j$ makes the sharding task more challenging since the tables will have more diverse dimensions and larger sizes. Given the number of GPUs $d$ and max dimension $2^j$, we sample a sharding task by randomly choosing $T$ tables from the table pool, where $10 \le T \le 60$ for 4 GPUs and $20 \le T \le 120$ for 8 GPUs. For each pair of $d$ and $2^j$, we randomly construct 100 sharding tasks. We provide a detailed description of the sharding tasks generation in Appendix~\ref{appendix:D}. In Section~\ref{sec:41}, we consider all the above $d$-$2^j$ pairs. For the other experiments, we mainly focus on a maximum dimension of 128 and 4 GPUs. Also, we consider a real-world sharding task in a production model, which aims to shard hundreds of tables to 128 GPUs (Section~\ref{sec:45}).

\begin{figure*}[t]
  \centering
  \begin{subfigure}[b]{0.33\textwidth}
    \centering
    \includegraphics[width=0.8\textwidth]{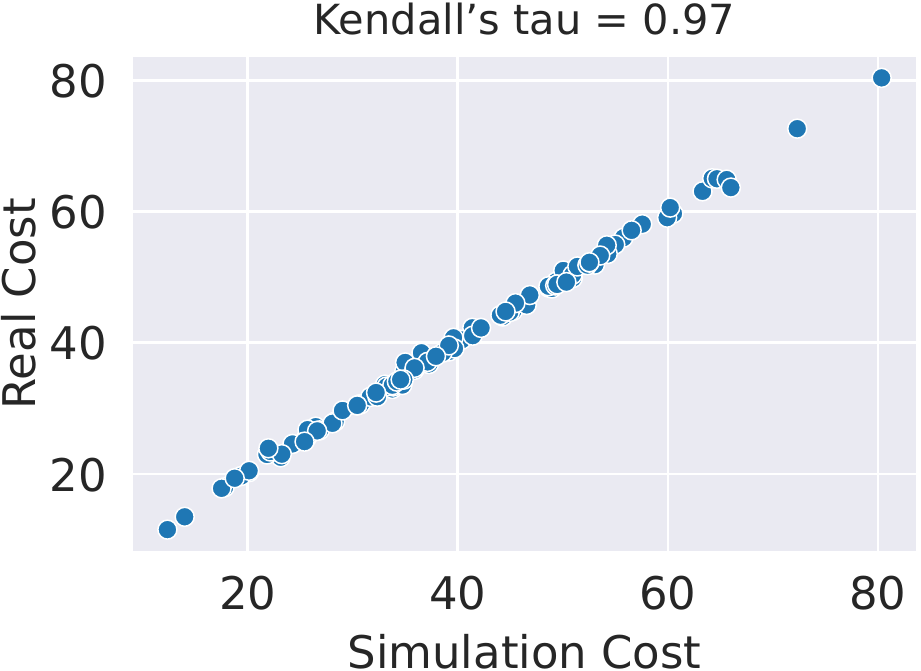}
   \end{subfigure}%
  \begin{subfigure}[b]{0.33\textwidth}
    \centering
    \includegraphics[width=0.8\textwidth]{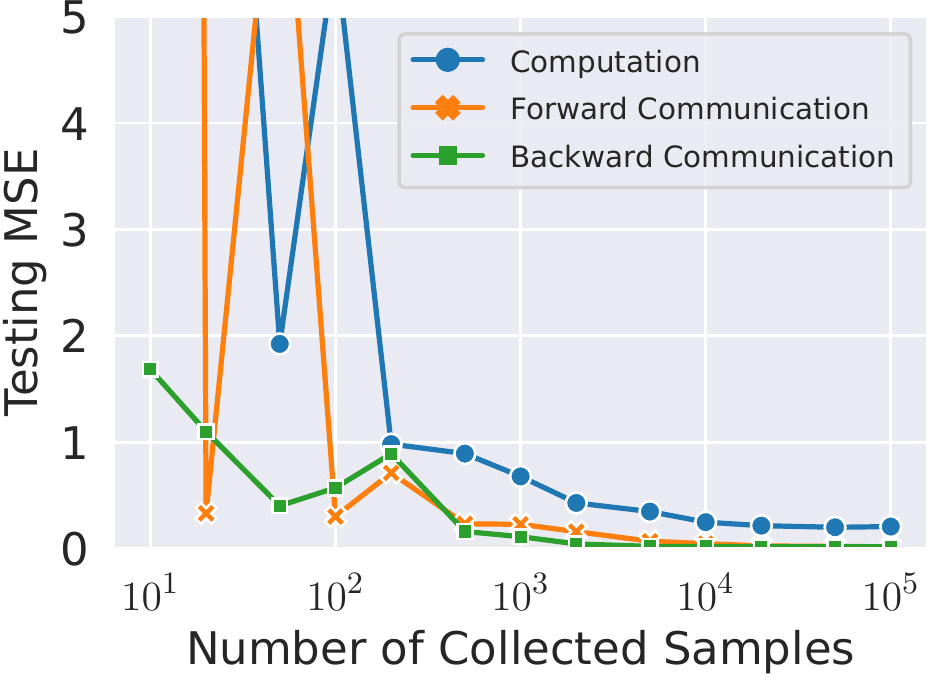}
   \end{subfigure}%
  \begin{subfigure}[b]{0.33\textwidth}
    \centering
    \includegraphics[width=0.8\textwidth]{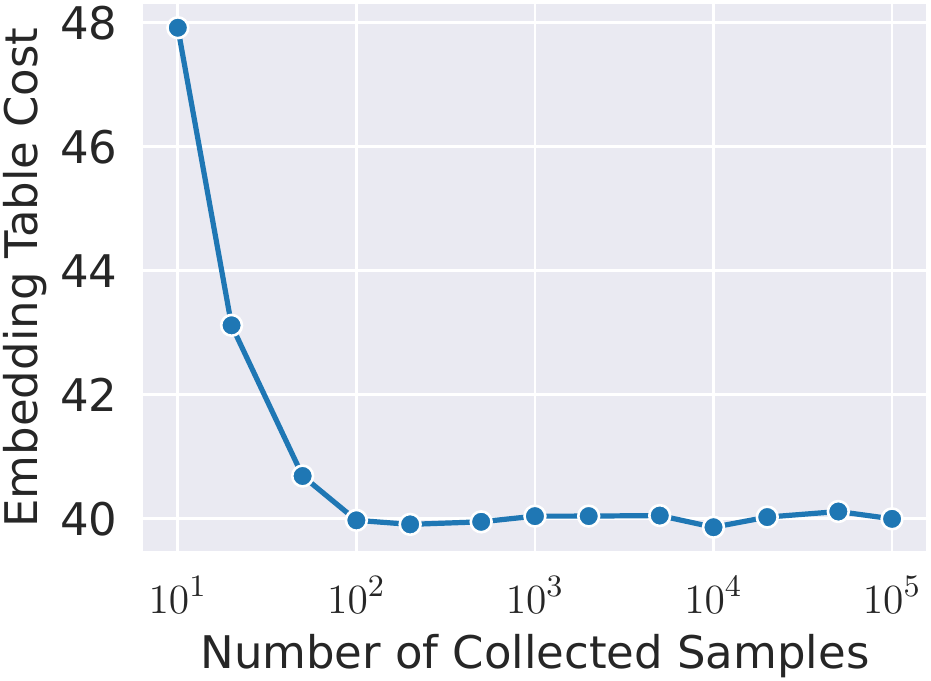}
   \end{subfigure}%
  \caption{Left: the scatter plot of the simulation costs estimated by the neural cost models and the real costs measured on GPUs. Middle: the performances of the neural cost models w.r.t. the number of samples used in training. Right: the embedding table cost for  the sharding tasks with a maximum dimension of 128 and 4 GPUs using the neural cost models that are trained with different numbers of samples. }
  \label{fig:costmodelimpact}
\end{figure*}

\textbf{Baselines.} We consider baselines in several categories (detailed in Appendix~\ref{appendix:E}): 1) \textbf{Random} sharding, 2) \textbf{Greedy} algorithms that balance various heuristic costs~\cite{acun2021understanding,lui2021understanding}, 3) \textbf{Reinforcmeent Learning} algorithms proposed in~\cite{zha2022autoshard,zha2022dreamshard}, and 4) \textbf{Planning} algorithm provided in TorchRec\footnote{\url{https://github.com/pytorch/torchrec}}.

\textbf{Evaluation protocol.} For each pair of $d$ and $2^j$, we apply each sharding algorithm to generate sharding plans for the 100 sharding tasks and collect real embedding costs from GPUs. To collect the costs, we run the embedding operations on GPUs to simulate computation and communication and use a timer to measure the time spent on each device. We report the maximum cost across devices for each sharding task since the maximum embedding cost will become the bottleneck. If any of the sharding plans generated by an algorithm causes memory error, it means the algorithm cannot scale to the setting defined by $d$ and $2^j$, so we denote the performance as ``-". If all the sharding plans are valid, we report the mean embedding cost across the 100 tasks. For the real-world production sharding task, we report both embedding costs and end-to-end training throughput improvements. The embedding costs are directly obtained from the traces collected during model training.

\textbf{Implementation details.} For the generation of the synthetic input, we augment the table pool with dimensions $\{4, 8, 16, 32, 64, 128\}$. We randomly select 1 to 15 tables for the table combination generation and $N_\text{place}$ tables for the table placement generation, where $10 \le N_\text{place} \le 60$ for 4 GPUs and $20 \le N_\text{place} \le 120$ for 8 GPUs. We generate 100K samples for each cost model. We randomly select a starting timestamp from 0 to 20 milliseconds. For the online search, we set $N=10$, $K=3$, $L=10$, and $M=11$. All the experiments are conducted on a server with eight 2080Ti GPUs. We provide more details in Appendix~\ref{appendix:E}.


\subsection{Comparison with the State-of-the-art Methods}
\label{sec:41}

To answer \textbf{RQ1}, we compare NeuroShard with the baselines on different numbers of GPUs and max table dimensions in Table~\ref{tab:performance}. We make the following observations. \textbf{1)} NeuroShard significantly outperforms the baselines in all the sharding tasks with an improvement ranging from +0.5\% to +23.8\%, which demonstrates the superiority of NeuroShard. An advantage of NeuroShard over AutoShard and DreamShard stems from its column-wise sharding. The large or costly tables will be partitioned to avoid memory explosion or become the bottleneck in sharding. This can not be achieved by DreamShard and AutoShard. \textbf{2)} NeuroShard successfully scales to sharding tasks with high table dimensions. When the table dimension goes large, all the baselines except TorchRec tend to fail. This is because it becomes harder to find a sharding plan that can satisfy the memory constraint with larger tables. Unlike the greedy and RL-based methods, NeuroShard searches for the best column-wise sharding plan so that it can partition the large tables. Surprisingly, the RL-based methods fail even when the dimension is small. A possible reason is that the stochastic policies in RL are very hard to train with high variance. \textbf{3)} Learning-based methods tend to perform better than heuristic costs. This is because the neural cost models are trained in a data-driven manner so they can provide a better cost estimation to boost the sharding performance. \textbf{4)} While TorchRec also scales well, NeuroShard achieves much better performances. This is because TorchRec still relies on a heuristic cost function, which is inaccurate. Whereas the cost models in NeuroShard are pre-trained in a data-driven manner so that they can estimate table costs more accurately.

\subsection{Analysis of Neural Cost Models}
\label{sec:42}
To study \textbf{RQ2}, we design experiments to understand how accurate the cost models are and how accurate they need to be to enable good sharding plans. First, we report the testing mean-squared-error (MSE) losses of all the pre-trained neural costs models in Table~\ref{tab:costmodel}. We observe that the largest MSE is 0.26, which suggests that the prediction error is within 0.6 milliseconds (0.6 $\times$ 0.6 = 0.36 $>$ 0.26). Note that the real costs may have some variance when collecting them. Thus, an error of 0.6 milliseconds is highly accurate. The left-hand side of Figure~\ref{fig:costmodelimpact} plots the real costs and simulation costs for 100 random sharding plans. The results again verify the high accuracy of the cost models.

Then, we visualize how many samples are needed to train the cost models in the middle and right-hand side of Figure~\ref{fig:costmodelimpact}. As expected, all the cost models become more accurate when we have more samples. Interestingly, even with only $10^2$ samples, NeuroShard can achieve very strong performance. This suggests that we only need sufficiently but not perfectly accurate cost models. This is desirable in practical use since it means we do not need many samples. Note that the result does not imply that we can use a simpler model. The current neural architecture of NeuroShard is already very shallow. An even simpler network (i.e., a linear one) may not work due to the non-linearity of the costs.

\begin{figure*}[t]
  \centering
  \begin{subfigure}[b]{0.7\textwidth}
    \centering
    \includegraphics[width=0.95\textwidth]{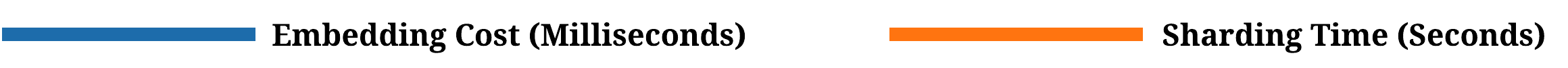}
   \end{subfigure}%

  \begin{subfigure}[b]{0.25\textwidth}
    \centering
    \includegraphics[width=0.95\textwidth]{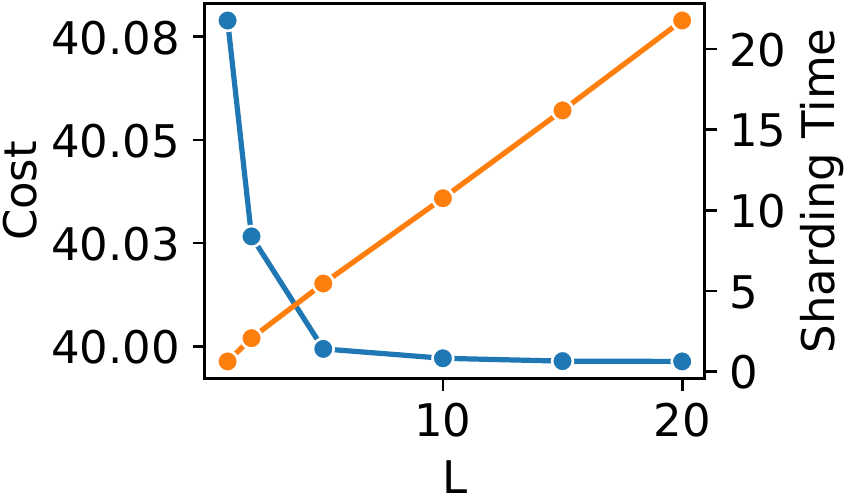}
   \end{subfigure}%
  \begin{subfigure}[b]{0.25\textwidth}
    \centering
    \includegraphics[width=0.95\textwidth]{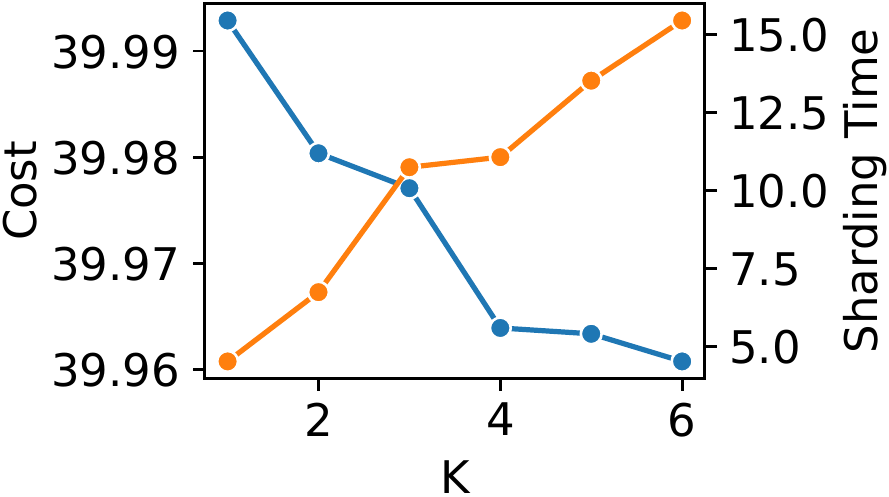}
   \end{subfigure}%
  \begin{subfigure}[b]{0.25\textwidth}
    \centering
    \includegraphics[width=0.95\textwidth]{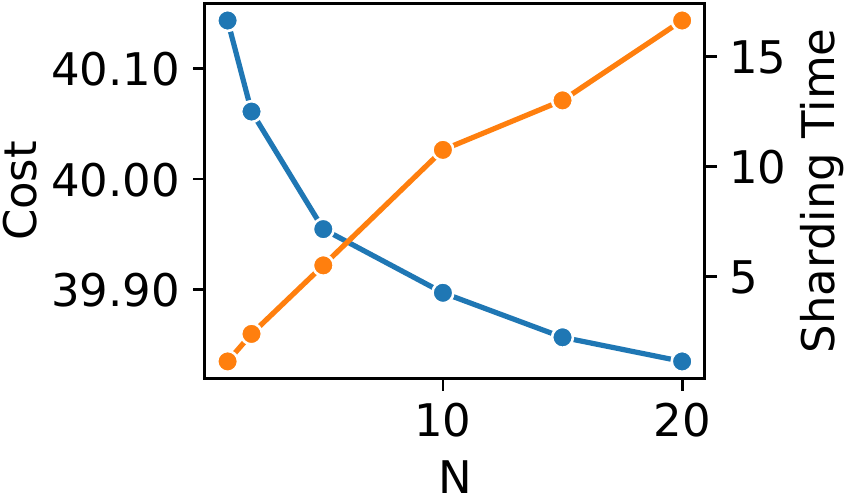}
   \end{subfigure}%
  \begin{subfigure}[b]{0.25\textwidth}
    \centering
    \includegraphics[width=0.95\textwidth]{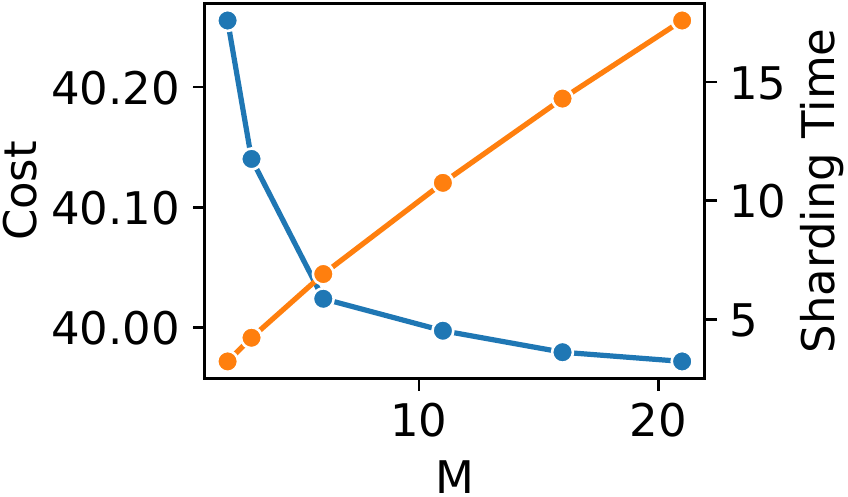}
   \end{subfigure}%
   \vspace{-8pt}
  \caption{Impact of NeuroShard's hyperparameters.}
  \vspace{-7pt}
  \label{fig:hyper}
\end{figure*}

\subsection{Ablation Study}
\label{sec:43}

To investigate \textbf{RQ3}, we report the results with one of the proposed beam search, greedy grid search, and caching removed in Table~\ref{tab:ablation}. We make two observations. \textbf{1)} The performance drops significantly when removing beam search or greedy grid search, which demonstrates the necessity of performing a joint search with both of them. \textbf{2)} The sharding takes significantly more time when removing the caching mechanism. This is because the cache has a more than 95\% hit rate, which can significantly accelerate the sharding speed.

\subsection{Hyperparamter Analysis}
\label{sec:44}

To understand \textbf{RQ4}, we analyze the impact of the hyperparameters in NeuroShard. Recall that we have 4 hyperparameters in the online search to balance between optimality and efficiency, i.e., $N$, $K$, $L$, and $M$. We visualize their impacts in Figure~\ref{fig:hyper}. We make two observations. \textbf{1)} A larger value for all the above four hyperparameters leads to better performance. This is because a larger value will result in more search iterations. \textbf{2)} Larger values also lead to more sharding time. Thus, we should specify an appropriate value (not too large nor too small) for each of the hyperparameters to strike a balance between optimality and efficiency.

\begin{table}[t]
    \centering
    \small
    \caption{Testing MSE of the neural cost models.}
    \label{tab:costmodel}
    \setlength{\tabcolsep}{3pt}
    \begin{tabular}{l|c|c|c}
    \toprule
     & DLRM & DLRM & Production \\
     & (4 GPUs) & (8 GPUs) & (128 GPUs) \\
    \midrule
     Computation & 0.21 & 0.21 & 0.26 \\
     Forward Communication & 0.02 & 0.05 & 0.05 \\
     Backward Communication & 0.02 & 0.04 & 0.15 \\
    \bottomrule
    \end{tabular}
\end{table}

\begin{table}[t]
    \centering
    \scriptsize
    \caption{Ablation study with a maximum dimension of 128 and 4 GPUs. w/o beam search means removing the beam search in col-wise sharding. w/o greedy grid search suggests not grid-searching the table dimension threshold. w/o caching disables the caching mechanism of computation costs. Results on 8 GPUs are in Appendix~\ref{appendix:G}.}
    \label{tab:ablation}
    \setlength{\tabcolsep}{0.4pt}
    \begin{tabular}{l|c|c|c|c}
    \toprule
     & Cost & \multirow{2}{*}{Success Rate} & Sharding Time & \multirow{2}{*}{Cache Hit Rate} \\
     & (Milliseconds) & ~ & (Seconds) \\
    \midrule
     w/o beam search & - & 87.0\%\ & 0.05 & 79.1\% \\
     w/o greedy grid search & 42.90 & 100.0\% & 2.12 & 82.2\% \\
     w/o caching & 39.99 & 100.0\% & 95.87 & 0.0\% \\
     \midrule
     Full NeuroShard & 39.99 & 100.0\% & 10.75 & 95.4\% \\
     
    \bottomrule
    \end{tabular}
\end{table}

\subsection{Application of NeuroShard to Production Models}
\label{sec:45}
To answer \textbf{RQ5}, we deploy NeuroShard to an ultra-large production DLRM. We used a state-of-the-art hardware platform with RDMA network fabrics, which is detailed in~\cite{mudigere2022software}. The model has nearly a thousand embedding tables that demand multi-terabyte memory. The task is to shard these tables to 128 GPUs. We compare NeuroShard with the baselines on embedding cost and training throughput in Table~\ref{tab:production}. Because we observed out-of-memory errors without a column-wise sharding, for the baselines except for TorchRec, we first apply the column-wise sharding plan proposed by NeuroShard and then run the baselines. Compared with the state-of-the-art (DreamShard), NeuroShard achieves 11.6\% improvement in embedding costs, which translates to 6.6\% end-to-end training throughput improvement. Note that $>$5\% is considered very significant in our production model since it has been heavily optimized. Also, NeuroShard can shard tables significantly faster in deployment; DreamShard requires RL training when applied to this model, while NeuroShard does not need further training since it directly leverages pre-trained cost models. Note that NeuroShard is trained using the data collected one month before the deployment, so there could be a distribution shift in table indices. The results suggest that we do not need to re-train NeuroShard for at least 1 month in production use.

In this experiment, the whole process (collecting data + training NeuroShard) takes roughly one hour with 128 GPUs, which is minor compared to the 6.6\% throughput improvement because 1) training a recommendation model can take up to a week so that NeuroShard can save several hours for each run, and 2) we do not need to frequently re-train NeuroShard, as discussed above.

\begin{table}[t]
\centering
\small
\caption{Embedding cost and overall training throughput improvement of NeuroShard and baselines on a production model. The results are collected from a training cluster with 128 GPUs.}
\label{tab:production}
\setlength{\tabcolsep}{4pt}
\begin{tabular}{l|c|c}
\toprule
\multirow{2}{*}{Sharding Algorithm} & Embedding Cost & Training Throughput \\
 ~ & (Milliseconds) & Improvement \\
\midrule
Random & 118.3 & - \\
Size-based & 107.6 & +4.0\%$\:\:$ \\
Dim-based & 90.8 & +13.9\% \\
Lookup-based & 102.4 & +11.9\% \\
Size-lookup-based & 109.2 & +12.8\% \\
AutoShard & 86.6 & +32.4\% \\
DreamShard & 61.6 & +45.3\%\\
TorchRec & 86.4 & +34.6\%\\
NeuroShard & 55.2 & +54.9\% \\

\bottomrule
\end{tabular}
\end{table}


\section{Related Work}

\textbf{DLRMs.} DLRMs have been widely adopted in many recommendation scenarios~\cite{zhang2019deep,cheng2016wide,naumov2019deep,he2017neural,wang2020skewness,lin2019negative,chuang2020tpr,chang2020query,zhou2021temporal,zhou2022multi,tan2021dynamic,tan2021sparse,tan2020learning,liu2019single,tan2019deep,tan2023s2gae}. To train DLRMs on ultra-large data and model sizes, distributed training solutions have been developed~\cite{acun2021understanding,covington2016deep,zhou2019deep,liu2017related,gomez2015netflix}. Embedding table sharding is an important design factor in the distributed training of DLRMs, which has been rarely studied in the literature. NeuroShard provides a deployable embedding table sharding solution to boost the distributed training efficiency of DLRMs.

\textbf{Embedding Table Sharding.} Several recent papers have studied embedding table sharding. The pioneering work relies on heuristic cost functions and greedy strategies for sharding~\cite{acun2021understanding,lui2021understanding}. RecShard formulates sharding as an optimization problem with mixed integer linear program~\cite{sethi2022recshard}. However, it does not consider the non-linearity of the table costs. FlexShard~\cite{sethi2023flexshard} presents a tailored sharding strategy for sequential DLRM. SurCo~\cite{ferber2022surco} solves embedding table sharding by learning linear Surrogate costs. Our previous work uses reinforcement learning (RL) to optimize sharding with learned cost models~\cite{zha2022autoshard,zha2022dreamshard}. While RL-based methods have achieved significant improvement, they cannot handle very large or costly tables, are expensive to train, and are unstable with high variance due to the stochastic policies in RL. In contrast, NeuroShard pre-trains cost models for a once-for-all purpose and jointly searches for column-wise and table-wise sharding plans. NeuroShard outperforms the RL-based methods in the benchmark sharding dataset and boosts the end-to-end training throughput of a production-scale model.

\textbf{Embedding Table Compression.} In parallel, researchers have studied how to compress embedding tables~\cite{zhang2020model,shi2020compositional,zhao2020autoemb,joglekar2020neural,liu2020learnable,kang2020learning,kang2021learning,pansare2022learning,desai2022random,lan2019albert,chen2015compressing}. Embedding table sharding is an orthogonal direction to these methods. This is because we often still need to perform table sharding after compression since the compressed tables can be still too large to fit on a single GPU's memory. Thus, sharding and compression can complement each other with their efficiency improvements. Moreover, embedding compression may lead to an accuracy drop since the embeddings may lose information. Whereas sharding is a lossless optimization on how to partition and place tables so it can improve efficiency without any accuracy loss.

\textbf{Device placement.} Another task that is related to embedding table sharding is the device placement of operations in the neural network. The existing work in this research line either uses reinforcement learning~\cite{mirhoseini2017device,mirhoseini2018hierarchical,gao2018spotlight,addanki2019placeto,paliwal2019reinforced,gao2018post,goldie2020placement} or cost modeling~\cite{lawler1993sequencing,jia2019beyond,jia2018exploring,narayanan2019pipedream,tarnawski2020efficient}. Our work is also based on cost modeling. Unlike regular operations, it is hard to estimate the cost of an embedding operation since the cost depends not only on the operation but also on the indices distributions. Our work presents a learning-based solution for embedding table sharding with pre-trained cost models and online search.

\section{Conclusion and Future Work}
In this work, we present NeuroShard, an embedding table sharding framework based on pre-trained neural cost models and online search. We have developed various strategies to train universal and accurate cost models for estimating embedding costs. With the pre-trained cost models, the online search requires minimum computational resources without real GPU execution. We show that NeuroShard not only outperforms the existing sharding algorithms on the benchmark dataset but also significantly boosts the training throughput of an ultra-large production DLRM. In the future, we will extend NeuroShard to row-wise sharding for partitioning large tables. Also, we plan to investigate CPU sharding or mixed CPU-GPU sharding scenarios. Last, we will explore \emph{``pre-train, and search''} for other system problems.

\vspace{-3pt}

\section*{Acknowledgements}
Rice University authors are, in part, supported by NSF (\#IIS-2224843). The views are those of the authors and should not be interpreted as representing any funding agencies. We'd like to thank the helpful feedback from the anonymous reviewers and Ankur Mallick during the shepherding process.


\bibliography{ref}
\bibliographystyle{mlsys2023}

\clearpage
\appendix

\section{Details of Analytical Experiments}
\label{appendix:A}

In this section, we provide more details on the three analytical experiments performed in Section~\ref{sec:2}.

\subsection{Impact of Dimension}
\label{appendix:A1}
\begin{figure*}[ht!]
  \centering
  \begin{subfigure}[b]{0.2\textwidth}
    \centering
    \includegraphics[width=0.95\textwidth]{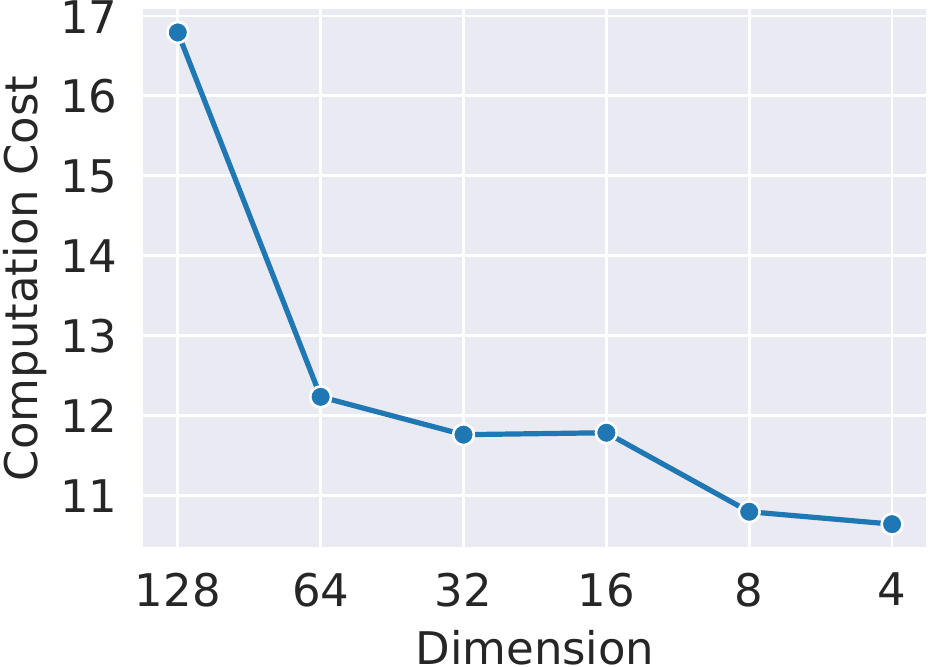}
  \end{subfigure}%
  \begin{subfigure}[b]{0.2\textwidth}
    \centering
    \includegraphics[width=0.95\textwidth]{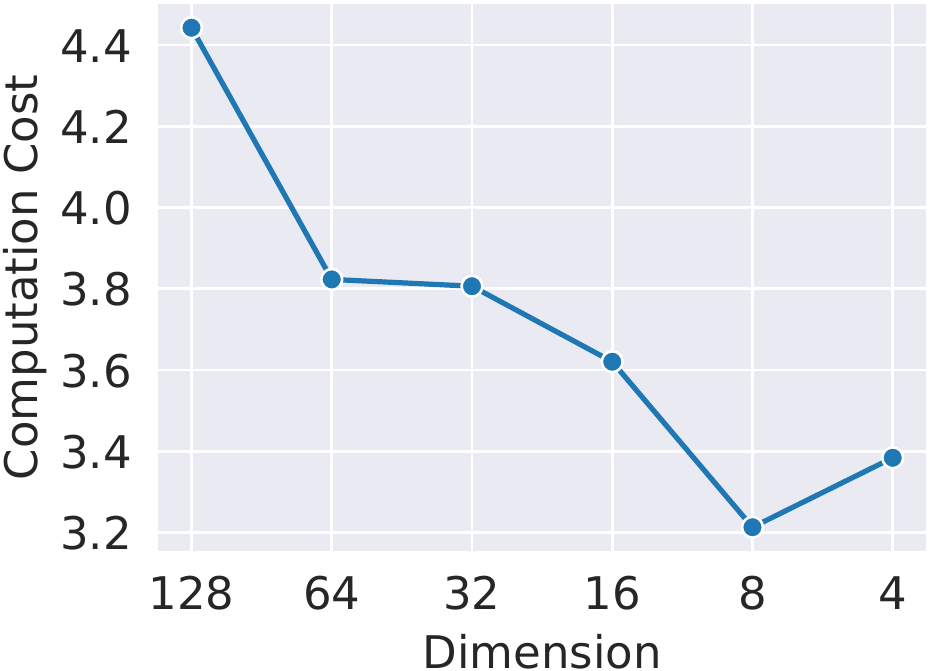}
  \end{subfigure}%
  \begin{subfigure}[b]{0.2\textwidth}
    \centering
    \includegraphics[width=0.95\textwidth]{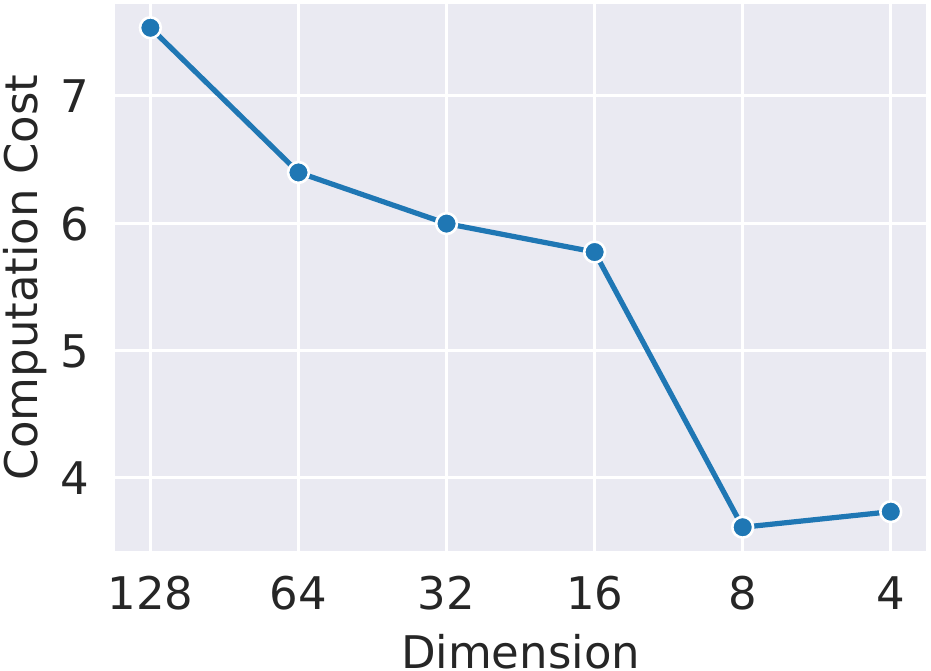}
  \end{subfigure}%
  \begin{subfigure}[b]{0.2\textwidth}
    \centering
    \includegraphics[width=0.95\textwidth]{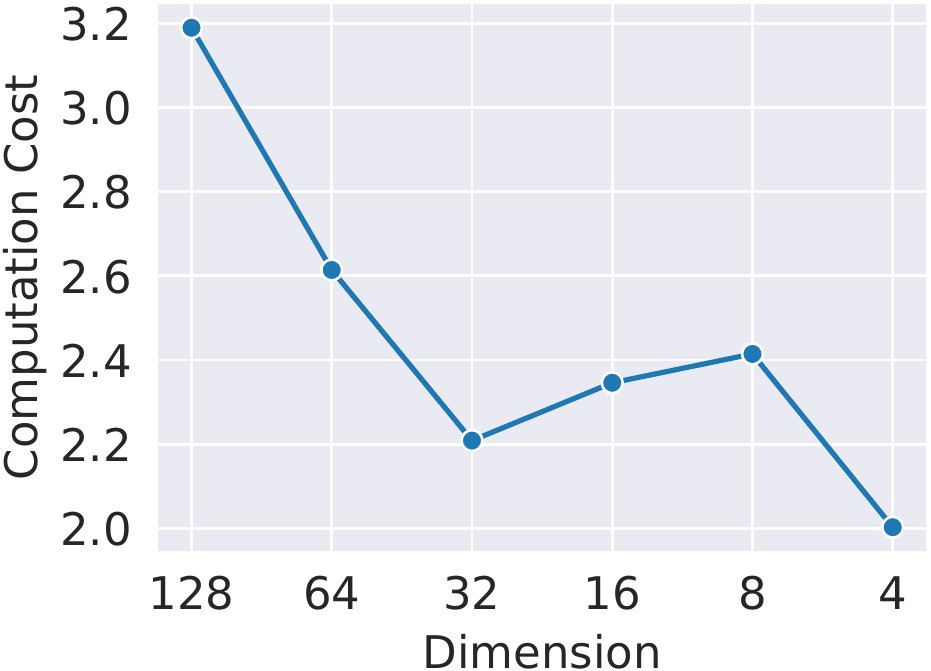}
  \end{subfigure}%
  \begin{subfigure}[b]{0.2\textwidth}
    \centering
    \includegraphics[width=0.95\textwidth]{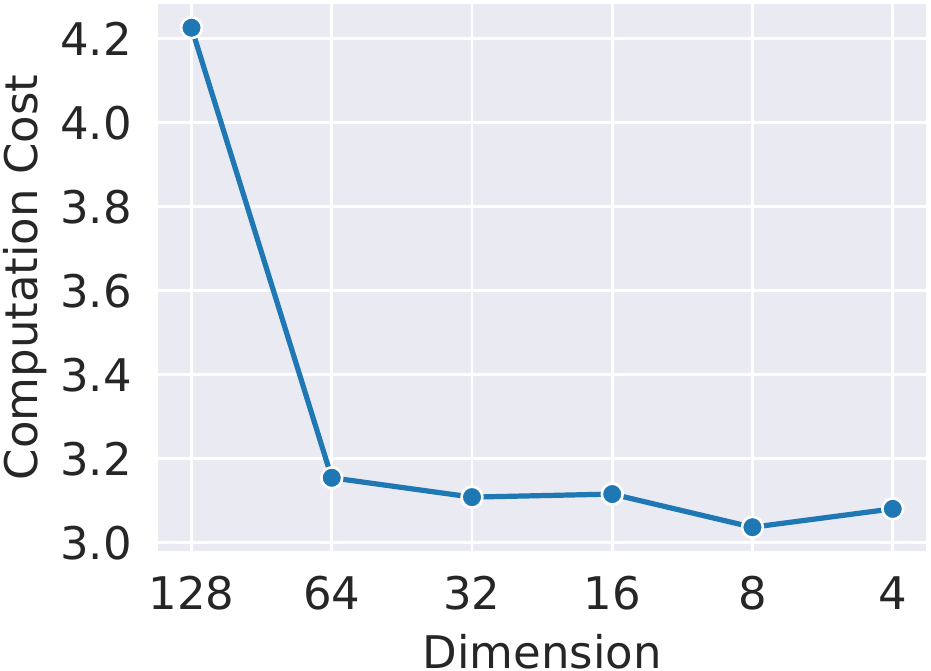}
  \end{subfigure}%
  \caption{Computation costs w.r.t. dimensions for some other randomly selected tables.}
  \label{fig:A1}
\end{figure*}

In this experiment, we aim to study how the dimension impacts the computation costs of the tables. In the DLRM dataset~\cite{naumov2019deep}, we have the table indices and hash sizes of the tables. The dimensions are not specified. Thus, we vary the dimension from the set of $\{128, 64, 32, 8, 4\}$ to study the influence of dimension. Note that for all the above five cases, we use the same hash size and the same table indices, and the only difference is the dimension. We have tried multiple tables and observed similar patterns. The results for some other randomly selected tables are visualized in Figure~\ref{fig:A1}. 

\subsection{Multi-table Costs vs the Sum of Single-table Costs}
\label{appendix:A2}
This analytical experiment aims to reproduce the results in~\cite{zha2022autoshard} to understand the relationship between the multi-table costs and the sums of the single-table costs. We randomly sample 50 data points to plot the scatters, where each data point consists of 10 tables from the 856 tables in the DLRM dataset~\cite{naumov2019deep}. For each of the data points, we measure two costs on a single GPU:
\begin{itemize}
    \item \textbf{Multi-table cost:} we directly run the fused operation on GPU (forward and backward passes). We first run the operation 10 times to warm up the hardware. Then we run another 100 times and use the median cost as the multi-table cost. 
    \item \textbf{Sum of single-table costs:} we collect the cost for every single table following the same process. Then we sum the single-table costs.
\end{itemize}
We observe that the sum of single-table costs is often larger than the multi-table costs because the fused operation is faster.

\subsection{Communication Cost Analysis}
\label{appendix:A3}
This analytical experiment aims to provide an understanding of the relationship between the max device dimension and the max forward/backward communication cost. To simulate the scenarios with different max device dimensions, we use the same random table placement generation strategy described in Section~\ref{sec:31}, with a greedy strategy to balance the dimensions equipped with randomness to cover different scenarios. A complete generation process is summarized in Algorithm~\ref{alg:placement}. For each data point (i.e., a placement), we run the forward all-to-all communication and the backward communication to collect the costs from the GPUs in a single server. Similar to computation costs, we first run the communication 10 times to warm up the hardware. Then we run the communication another 100 times and use the median costs as the communication costs. Note that each GPU will have a different locally measured cost. we only focus on the max cost since it is the bottleneck.

\section{Details of Synthetic Inputs Generation}
\label{appendix:B}
In this section, we give more details about how we generate synthetic data to have good coverage for benchmarking.

\subsection{Table Augmentation}
\label{appendix:B1}
The key idea is to generate various dimensions for a single table. Given a list of dimensions, for each table in the pool, we associate the table with each of the dimensions as an augmented table. Algorithm~\ref{alg:augmentation} summarizes the augmentation procedure.

\begin{algorithm}[ht!]
\caption{Table Augmentation}
\label{alg:augmentation}
\setlength{\intextsep}{0pt} 
\begin{algorithmic}[1]
\STATE \textbf{Input:} $T$ embedding tables, a set of dimensions $\mathcal{D}$
\STATE Augmented table pool $\mathcal{P}$ $\leftarrow \{\}$
\FOR{each of the $T$ tables}
    \FOR{each of the dimensions in $\mathcal{D}$}
        \STATE The augmented table $aug\_table \leftarrow$ the selected table with the selected dimension
        \STATE Add $aug\_table$ to $\mathcal{P}$
    \ENDFOR
\ENDFOR
\STATE Return $\mathcal{P}$

\end{algorithmic}
\end{algorithm}

\subsection{Random Table Combination Generation}
\label{appendix:B2}
We summarize the process of generating random data table combinations in Algorithm~\ref{alg:combination}.

\begin{algorithm}[ht!]
\caption{Random Table Combination Generation}
\label{alg:combination}
\setlength{\intextsep}{0pt} 
\begin{algorithmic}[1]
\STATE \textbf{Input:} Augmented table pool $\mathcal{P}$, min number of table $T_{\text{min}}$, max number of table $T_{\text{max}}$, number of table combinations we aim to generate $N_{\text{com}}$
\STATE Table combinations $\mathcal{T}_{\text{com}}$ $\leftarrow \{\}$
\FOR{$i=0, 1, ..., N_{\text{com}}$}
    \STATE Uniformly sample the number of tables $T$ in $[T_{\text{min}}, T_{\text{max}}]$
    \STATE Randomly sample $T$ tables from $\mathcal{P}$ and append this combination to $\mathcal{T}_{\text{com}}$
\ENDFOR
\STATE Return $\mathcal{T}_{\text{com}}$

\end{algorithmic}
\end{algorithm}

\subsection{Random Table Placement Generation}
\label{appendix:B3}
We summarize the table placement generation in Algorithm~\ref{alg:placement}.

\begin{algorithm}[ht!]
\caption{Random Table Placement Generation}
\label{alg:placement}
\setlength{\intextsep}{0pt} 
\begin{algorithmic}[1]
\STATE \textbf{Input:} Augmented table pool $\mathcal{P}$, min number of table $T_{\text{min}}$, max number of table $T_{\text{max}}$, number of table placements we aim to generate $N_{\text{place}}$, number of GPU devices $D$.
\STATE Table placements $\mathcal{T}_{\text{place}}$ $\leftarrow \{\}$
\FOR{$i=0, 1, ..., N_{\text{place}}$}
    \STATE Uniformly sample the number of tables $T$ in $[T_{\text{min}}, T_{\text{max}}]$
    \STATE Randomly sample $T$ tables from $\mathcal{P}$ 
    \STATE Sort the $T$ tables in descending order based on the table dimension
    \STATE Uniformly sample a probability of applying greedy strategy $p \in [0, 1]$
    \FOR{each of the $T$ tables}
        \STATE Randomly sample $p' \in [0, 1]$
        \STATE Obtain the candidate GPU devices that will not cause a memory error.
        \IF{$p' \le p$}
            \STATE Assign the current table to the candidate GPU with the lowest device dimension 
        \ELSE  
            \STATE Randomly assign the current table to one of the candidate GPUs
        \ENDIF
    \ENDFOR
    \STATE Append the placement to $\mathcal{T}_{\text{place}}$
\ENDFOR
\STATE Return $\mathcal{T}_{\text{place}}$

\end{algorithmic}
\end{algorithm}

\section{Details of Cost Model Training}
\label{appendix:C}
In this section, we provide more details of the neural architecture and loss functions for training the neural cost models.

\textbf{Neural architectures.} For the computation cost model, we use an MLP with a size of 128-32 to process the table features and another MLP with a size of 32-64. For the communication cost model, we use an MLP with a size of 128-64-32-16.

\textbf{Loss functions.} We use mean squared error (MSE) to update both the computation cost model and communication cost model. Specifically, let $\mathbf{x}$ be the features (can be either table features or communication features), $y$ be the ground truth, and $g$ be the cost model (can be either computation cost model or communication model). Let $N_{\text{sample}}$ be the number of samples. Then the loss is

\begin{equation}
    \mathcal{L} = \sum_{i=1}^{N_{\text{sample}}} \text{MSE}(\mathbf{x}_i, y_i),
\end{equation}
where $\text{MSE}(\cdot, \cdot)$ represents the MSE loss. In practice, we can use mini-batch training.

\section{Datasets}
\label{appendix:D}

We use the benchmark dataset for evaluating embedding table sharding~\cite{naumov2019deep}. It is publicly available as  \url{https://github.com/facebookresearch/dlrm_datasets}. It contains 856 synthetic tables whose indices distributions are similar to the production workloads in Meta. The statistics of the datasets are well summarized in previous work. Please see the appendices in~\cite{zha2022autoshard,zha2022dreamshard} for details.

We summarize the 12 sharding tasks we used in our experiments in Table~\ref{tab:shardingtasks}. All the sharding tasks have a constraint of 4 GB memory in each GPU.

\begin{table*}[ht!]
    \centering
    \caption{Sharding tasks generated in the experiments.}
    \label{tab:shardingtasks}
    \setlength{\tabcolsep}{2pt}
    \begin{tabular}{l|l|l}
    \toprule
    Number of GPUs & Range of the Number of Tables & Range of Table Dimensions \\
    \midrule
     4 & 10-60 & 4 \\
     4 & 10-60 & 4, 8 \\
     4 & 10-60 & 4, 8, 16 \\
     4 & 10-60 & 4, 8, 16, 32 \\
     4 & 10-60 & 4, 8, 16, 64 \\
     4 & 10-60 & 4, 8, 16, 64, 128 \\
     8 & 20-120 & 4 \\
     8 & 20-120 & 4, 8 \\
     8 & 20-120 & 4, 8, 16 \\
     8 & 20-120 & 4, 8, 16, 32 \\
     8 & 20-120 & 4, 8, 16, 64 \\
     8 & 20-120 & 4, 8, 16, 64, 128 \\
     
    \bottomrule
    \end{tabular}
\end{table*}

\textbf{Discussion of public datasets.} Table~\ref{tbl:datascale} compares the scale of DLRM datasets with several large-scale public datasets. DLRM has significantly more tables, a larger average hash size, and a larger average pooling factor than these public datasets. The embedding costs on these datasets will always be very small, no matter how we do sharding. For example, DLRM has at least 30x more tables than Criteo, at least 200x larger average hash size, and a 15x larger average pooling factor. So, as the embedding cost of the DLRM dataset is from the range of 17 ms to 40 ms (Table~\ref{tab:performance} in our paper), the embedding cost of Criteo could be only around 1 ms or even smaller. Thus, there is no need to do sharding on the Criteo dataset. So these datasets are not suitable for evaluating embedding table sharding algorithms.

\begin{table*}[ht!]
\centering
\caption{Comparison of embedding table feature statistics between some popular public recommendation datasets and the industrial-scale DLRM dataset. }
\label{tbl:datascale}
\scriptsize

\begin{tabular}{l|l|c|c|c|l}
\toprule
\multicolumn{2}{c|}{Dataset} & \# of Tables & Avg. hash size  & Avg. pooling factor & Link \\
\midrule
\multirow{3}{*}{Public} & Criteo & 26 & 17,839 & 1 & \url{https://www.kaggle.com/c/criteo-display-ad-challenge} \\
~ & Avazu & 23 & 67,152 & 1 & \url{https://www.kaggle.com/c/avazu-ctr-prediction/data} \\
~ & KDD & 10 & 601,908 & 1 & \url{https://www.kaggle.com/c/kddcup2012-track2/data} \\

\midrule
\multirow{1}{*}{Industrial-Scale} & DLRM & 856 & 4,107,458 & 15 & \url{https://github.com/facebookresearch/dlrm_datasets} \\
\bottomrule
\end{tabular}
\end{table*}

\section{Baselines}
\label{appendix:E}
In this section, we introduce the details of all the baselines used in our experiments.

\subsection{Greedy Algorithms}

The greedy sharding algorithms have been used in previous papers of distributed recommender systems~\cite{acun2021understanding,lui2021understanding}. The main idea is to use a greedy algorithm to balance the costs by assigning the table to the device with the lowest cost so far in each step, where the costs are estimated in different ways. Specifically, greedy algorithms consist of two steps as follows.

\begin{itemize}
    \item \textbf{Designing a cost function:} we give each table a cost to quantify the expected running time on the device. The cost is the objective that we want to balance.
    \item \textbf{Greedy allocation:} the objective is to balance the sum of the costs in each device. To achieve this, we first sort the embedding tables in descending order based on the costs defined by the cost function. The sorting can make it more easily to achieve a balance if we allocate the tables greedily. Then, we assign tables starting from the table with the highest cost. In each step, we make a greedy decision by assigning the current table to the device that has the lowest sum of the cost so far. This sorting-enhanced greedy strategy can enable each device to have roughly a similar sum of the costs.
\end{itemize}
The four baselines differ in how the cost function is designed. This can significantly impact the balance since the cost function determines our optimization objective. We summarize the used cost functions as follows:
\begin{itemize}
    \item \textbf{Size-based:} we use the table size as the cost function. The idea is that balancing table size can reduce the risk of getting out-of-memory errors. Also, table size is also positively correlated with dimension so it can also reflect the workloads.
    \item \textbf{Dim-based:} we use the table dimension as the cost function. Table dimension is an important feature to represent the cost since it can decide both computation and communication workloads. Thus, it is natural to balance the sums of dimensions (i.e., the device dimension).
    \item \textbf{Lookup-based:} we use the product of the table dimension and the mean pooling factor as the cost function. The intuition is that the table dimension and the pooling factor can determine the computation workload in embedding lookup.
    \item \textbf{Size-lookup-based:} we use the product of the table dimension, the mean pooling factor, and the table size as the cost function. This is a more comprehensive cost function that considers both lookup cost and table sizes.
\end{itemize}

\subsection{Reinforcement Learning Algorithms}
We have included two state-of-the-art reinforcement learning algorithms for embedding table sharding~\cite{zha2022autoshard,zha2022dreamshard}. They share similar ideas with differences in optimization objectives and training methods. We summarize them as follows.
\begin{itemize}
    \item \textbf{AutoShard}~\cite{zha2022autoshard}: it trains an LSTM controller to perform sharding to balance computation cost. The objective is the degree of balance, which is defined as the min cost divided by the max cost. The code is available as \url{https://github.com/daochenzha/autoshard}
    \item \textbf{DreamShard}~\cite{zha2022dreamshard}: it extends AutoShard by also balancing communication. It also extends the cost model to communication. It additionally introduces an estimated MDP to make training and inference much faster. The code is available at \url{https://github.com/daochenzha/dreamshard}
\end{itemize}

\subsection{Planing Algorithms}
In parallel to reinforcement learning, planning algorithms identify the sharding plan with search. TorchRec provides a planning-based sharding strategy. For a fair comparison, we allow TorchRec to search for both column-wise and table-wise sharding plans. However, TorchRec still relies on heuristic costs so we do not see a clear improvement of TorchRec over the greedy algorithms. The code is publicly available at \url{https://github.com/pytorch/torchrec}.

\section{Hyperprameters And Configurations}
\label{appendix:F}
In this section, we list all the hyperparameters of NeuroShard. We also list the hardware/software configurations

\begin{itemize}
    \item \textbf{Generating synthetic inputs:} we augment the table pool with dimensions $\{4, 8, 16, 32, 64, 128\}$. We randomly select 1 to 15 tables for the table combination generation and $N_\text{place}$ tables for the table placement generation, where $10 \le N_\text{place} \le 60$ for 4 GPUs and $20 \le N_\text{place} \le 120$ for 8 GPUs. The starting timestamp for the communication data is sampled from 0 to 20 milliseconds. We generate 100K samples for each of the cost models.
    \item \textbf{Training neural cost models:} We use 80\% of the data for training, 10\% of the data for validation, and 10\% of the data for testing. We set the batch size to 512. We use Adam optimizer with a learning rate of 0.001 with the other configurations as the default. We train 1000 epochs and save the model that can deliver the best results on the validation data.
    \item \textbf{Online search:} we set $N=10$, $K=3$, $L=10$, and $M=11$.
    \item \textbf{Software:} we use PyTorch 1.9.1, and FBGEMM 0.0.1
    \item \textbf{Hardware:} we conduct the experiments on a server with 48 Intel(R) Xeon(R) Silver 4116 CPU @ 2.10GHz processors, 188 GB memory, and eight NVIDIA GeForce RTX 2080 Ti GPUs.
\end{itemize}

\section{Additional Ablation Results}
\label{appendix:G}

The results are shown in Table~\ref{tab:ablation2}.

\begin{table*}[ht!]
    \centering
    \caption{Ablation study with a maximum dimension of 128 and 8 GPUs. w/o beam search means removing the beam search in col-wise sharding. w/o greedy grid search suggests not grid-searching the table dimension threshold. w/o caching disables the caching mechanism of computation costs.}
    \label{tab:ablation2}
    \setlength{\tabcolsep}{2pt}
    \begin{tabular}{l|c|c|c|c}
    \toprule
     & Cost & \multirow{2}{*}{Success Rate} & Sharding Time & \multirow{2}{*}{Cache Hit Rate} \\
      & (Milliseconds) & ~ & (Seconds) \\
    \midrule
     w/o beam search & - & 63.0\% & 0.09 & 75.4\% \\
     w/o greedy grid search & 55.97 & 100.0\% & 4.09 & 79.8\% \\
     w/o caching & 49.10 & 100.0\% & 164.21 & 0.0\% \\
     \midrule
     Full NeuroShard & 49.10 & 100.0\% & 20.79 & 93.0\% \\
     
    \bottomrule
    \end{tabular}
\end{table*}

\section{Additional Discussion of Search with Reinforcement Learning}

In this work, we used beam search and greedy grid search to identify in search. However, this search process could be further accelerated by training a meta-policy~\cite{zha2020meta,lai2020policy} with reinforcement learning~\cite{zha2021rlcard} and transferring it across tasks. Here, we highlight several potential strategies that we are trying. \textbf{1) Hierarchical reinforcement learning:} The idea is to decompose the sharding task into several sub-tasks~\cite{kulkarni2016hierarchical,zha2022towards}. This can be naturally applied to the embedding table sharding problem, as there are two hierarchies of search: column-wise sharding and table-wise sharding. \textbf{2) Self-imitation learning:} The idea is to select the highly-rewarded samples and use supervise losses to encourage the policy to reproduce the good behaviors~\cite{oh2018self,zharank,zha2021simplifying,li2021automated}. This could be helpful since we often have lots of system logs of sharding. The idea is to select good sharding plans from the system log and use supervised losses to train a policy. \textbf{2) Offline reinforcement learning:} The idea is to learn the optimal strategy based on offline data~\cite{kumar2020conservative,li2023towards}. This can also be applied to the offline sharding log.

Note that the reinforcement learning meta-policy could also be combined with search to guide the search process.

\section{Artifact}


This section is for readers who are interested in reproducing our results. In what follows, we will describe how to download the dataset, how to install NeuroShard and its dependencies, how to run NeuroShard, and how to collect latencies from the hardware.

\subsection{Artifact check-list (meta-information)}

{\small
\begin{itemize}
  \item {\bf Algorithm: }Pre-trained models, beam search, grid search.
  \item {\bf Program: }Implemented using PyTorch with Python.
  \item {\bf Compilation: }PyTorch 1.8.0 and Python 3.8.0.
  \item {\bf Data set: }Synthetic embedding data open-sourced by Meta.
  \item {\bf Run-time environment: }Ubuntu 18.04.6 LTS.
  \item {\bf Hardware: }Eight NVIDIA GeForce RTX 2080 Ti GPUs.
  \item {\bf Metrics: }Latency in milliseconds.
  \item {\bf Output: }Command line outputs.
  \item {\bf Experiments: }Generate sharding plans and collect latencies.
  \item {\bf How much disk space required (approximately)?: }20 GB.
  \item {\bf How much time is needed to prepare workflow (approximately)?: }2 hours.
  \item {\bf How much time is needed to complete experiments (approximately)?: }20+ hours.
  \item {\bf Publicly available?: } Yes.
  \item {\bf Code licenses (if publicly available)?: }MIT.
  \item {\bf Archived (provide DOI)?: }\url{https://zenodo.org/badge/latestdoi/556106683}
\end{itemize}}

\subsection{Description}

\subsubsection{How delivered}

The artifact is zipped and available at \url{https://zenodo.org/badge/latestdoi/556106683}. The source code is also publicly released on GitHub at \url{https://github.com/daochenzha/neuroshard}, with a README file that provides step-to-step instructions to run the code. 

\subsubsection{Hardware dependencies}
Our results are collected from a server with eight NVIDIA GeForce RTX 2080 Ti GPUs. Other types of GPUs are also acceptable but the results may vary since the collected latencies are GPU-dependent.

\subsubsection{Software dependencies}
To run the code, Python 3.8.0 or higher (we used 3.8.0) is required. Additionally, an appropriate version of CUDA must be installed to utilize GPUs. Lastly, FBGEMM, an open-sourced embedding table operation, must be installed.

\subsubsection{Data sets}
The synthetic dataset is publicly available at \url{https://github.com/facebookresearch/dlrm_datasets.git}. The dataset can be downloaded with \texttt{Git LFS} by running the following commands:

{\scriptsize
\begin{verbatim}
git lfs install --skip-smudge
git clone \
    https://github.com/facebookresearch/dlrm_datasets.git
cd dlrm_datasets
git lfs pull \
    --include=embedding_bag/2021/fbgemm_t856_bs65536.pt.gz
gzip -d embedding_bag/2021/fbgemm_t856_bs65536.pt.gz
\end{verbatim}
}

A file named \texttt{fbgemm\_t856\_bs65536.pt} will be obtained in \texttt{embedding\_bag/2021/} after running the above commands, and its size is 4.0 GB. This file contains synthetic indices for embedding lookups, which share similar indices distributions as the Meta production environment.

\subsection{Installation}

Firstly, we need to install Python 3.8.0+ and CUDA. Secondly, we install PyTorch 1.8.0 with the following:

{\scriptsize
\begin{verbatim}
pip3 install torch==1.8.0
\end{verbatim}
}

Thirdly, we need to install the open-sourced FBGEMM, which is available at \url{https://github.com/pytorch/FBGEMM/tree/main/fbgemm_gpu}. Note that, except for A100 or V100 GPUs, FBGEMM needs to be built manually following the instructions, which is expected to take 0.5 to 1 hour. Finally, clone the NeuroShard code and install it:

{\scriptsize
\begin{verbatim}
git clone \
    https://github.com/anonymoussubmition/neuroshard.git
cd neuroshard
pip3 install -e .
\end{verbatim}
}

\subsection{Experiment workflow}

The experiment consists of four main steps:
\begin{enumerate}
    \item We process the raw synthetic data with a script and randomly construct some sharding tasks for evaluation.
    \item We do micro-benchmarking on hardware to collect computation and communication costs. These cost data will be saved in a file.
    \item Based on the collected cost data, we pre-train cost models. Specifically, we train three models, including a computation cost model, a forward communication cost model, and a backward communication cost model.
    \item Using the pre-trained cost models, we can run the online search in NeuroShard for embedding table sharding. We can also run the baseline heuristic sharding algorithms. The performance can be evaluated by using the cost models (simulation) or collecting costs from the hardware.
\end{enumerate}
Now we walk through the above steps one by one. To begin with, we need to ensure that the current directory is in \texttt{neuroshard/} and put the \texttt{fbgemm\_t856\_bs65536.pt} file in the current directory. \textbf{In Step 1}, we run the following:

{\scriptsize
\begin{verbatim}
python3 tools/gen_dlrm_data.py \
  --data fbgemm_t856_bs65536.pt
\end{verbatim}
}

Here, \texttt{--data} specifies the path of the raw data. The expected output is:

{\scriptsize
\begin{verbatim}
Processing DLRM data...
Generating table configs...
\end{verbatim}
}

The processed data will be saved in \texttt{data/dlrm\_datasets} by default. After this, we generate sharding tasks with

{\scriptsize
\begin{verbatim}
python3 tools/gen_tasks.py --max_dim 128
\end{verbatim}
}

Here, \texttt{--max\_dim} specifies the maximum table dimension (the second row in Table 1). We used 128 as an example to show how to run the code. The expected output is: 

{\scriptsize
\begin{verbatim}
100 sharding tasks generated!
\end{verbatim}
}

The sharding tasks will be saved in \texttt{data/tasks/4\_gpus} by default. \textbf{In Step 2}, we collect cost data with micro-benchmarking. We first collect computation cost data:

{\scriptsize
\begin{verbatim}
python3 collect_compute_cost_data.py --data_size 10
\end{verbatim}
}

\texttt{--data\_size} specifies how many data samples will be generated. We use a small number here so that the program runs faster. To reproduce the results, \texttt{--data\_size} should be set as 100K. The expected final two lines should be similar to the following (the exact values in the first line may differ):

{\scriptsize
\begin{verbatim}
2997,436,2332,3728,535 3.4415176584173777
Device 0 finished!
\end{verbatim}
}

Similarly, we can collect communication cost data by running the following script:

{\scriptsize
\begin{verbatim}
python3 collect_comm_cost_data.py --data_size 10
\end{verbatim}
}

We also set \texttt{--data\_size} to be small here to make it run faster. 100K should be set instead to reproduce the results. The last line of the expected output should be:

{\scriptsize
\begin{verbatim}
Evaluator sub-process terminated!
\end{verbatim}
}

After running the above two scripts, the cost data is expected to be stored in \texttt{data/cost\_data/}. \textbf{In Step 3}, based on the collected cost data, we pre-train neural cost models. We train the computation cost model by running the following command:

{\scriptsize
\begin{verbatim}
python train_compute_cost_model.py --epochs 3
\end{verbatim}
}

\texttt{--epochs} means the number of epochs of training. We set it to be 3 here to make the training faster. To reproduce the result, \texttt{--epochs} needs to be set to 1000. The last line of the expected output is (the exact values may differ):

{\scriptsize
\begin{verbatim}
Final result, train MSE: 8.861547689180117,
valid MSE 8.398209571838379, test MSE: 8.432512283325195
\end{verbatim}
}

Then we do similar things again for training communication cost models. We can run the following command:

{\scriptsize
\begin{verbatim}
python3 train_comm_cost_model.py --epochs 3
\end{verbatim}
}

Again, we set \texttt{--epochs} to be 3 to make the training faster, and 1000 is needed to reproduce the results. The last line of the expected output is (the exact values may differ):

{\scriptsize
\begin{verbatim}
Final result, train MSE: 31.95864486694336,
valid MSE 42.04985046386719, test MSE: 23.29010772705078
\end{verbatim}
}

After training, there should be three models stored in \texttt{models/}.
\textbf{In Step 4}, we evaluate different sharding algorithms with simulation or with real hardware. To get simulation results, we can use the following commands to run NeuroShard and the heuristic baselines:

{\scriptsize
\begin{verbatim}
python3 eval_simulator.py --alg neuroshard
python3 eval_simulator.py --alg random
python3 eval_simulator.py --alg dim_greedy
python3 eval_simulator.py --alg lookup_greedy
python3 eval_simulator.py --alg size_greedy
python3 eval_simulator.py --alg size_lookup_greedy
\end{verbatim}
}

Similarly, we evaluate the cost on real hardware with

{\scriptsize
\begin{verbatim}
python3 eval.py --alg neuroshard
python3 eval.py --alg random
python3 eval.py --alg dim_greedy
python3 eval.py --alg lookup_greedy
python3 eval.py --alg size_greedy
python3 eval.py --alg size_lookup_greedy
\end{verbatim}
}

\subsection{Evaluation and expected result}

For each of the scripts in Step 4, the final result will be printed out in the terminal. The cost models need to be trained with \texttt{--data\_size} of 100K and \texttt{--epochs} of 3 in order to get similar results shown below. For the simulation result of NeuroShard, the expected output is:

{\scriptsize
\begin{verbatim}
Average: 39.0441405081749
Valid 100 / 100
\end{verbatim}
}

The first line is the average latency. The second line means NeuroShard succeeds on all the tasks, i.e., no memory error. If running \texttt{size\_lookup\_greedy}, the result will be:

{\scriptsize
\begin{verbatim}
Average: 47.95177095494372
Valid 94 / 100
\end{verbatim}
}

It only succeeds in 94 out of 100 tasks. For the results on real hardware, the expected output of NeuroShard is:

{\scriptsize
\begin{verbatim}
Average: 39.98647058823529
Valid 100 / 100
\end{verbatim}
}

The expected result of \texttt{size\_lookup\_greedy} is:

{\scriptsize
\begin{verbatim}
Average: 48.010588235294115
Valid 94 / 100
\end{verbatim}
}

The simulation and real costs are consistent. The above procedure only provides one column of the results in Table 1. To get full results, we need to customize the \texttt{--max\_dim}, \texttt{--T\_range}, and \texttt{--max\_mem} in \texttt{gen\_tasks.py}, and also the corresponding arguments in \texttt{collect\_comm\_cost\_data.py}, \texttt{eval.py}, and \texttt{eval\_simulator.py}.

\subsection{Notes}

The cost is highly dependent on the GPUs used, and different PyTorch versions may produce varying results. Additionally, a result itself may have a variance, so the results obtained on another machine may differ from those shown above.


\end{document}